\documentclass[10pt,twocolumn,letterpaper,dvipsnames,table]{article}

\usepackage[pagenumbers]{cvpr} 

\usepackage[dvipsnames]{xcolor}

\usepackage{cuted}

\definecolor{cvprblue}{rgb}{0.21,0.49,0.74}
\usepackage[pagebackref,breaklinks,colorlinks,citecolor=cvprblue]{hyperref}
\usepackage[capitalize]{cleveref}
\usepackage{bm}

\usepackage{adjustbox}
\usepackage{multirow}
\usepackage{graphicx}
\usepackage{xcolor}
\usepackage{booktabs}
\usepackage{caption}
\usepackage{makecell}
\usepackage{arydshln}

\def\eg{\emph{e.g.}\xspace}

\def\wo{\emph{w/o}\xspace}

\newcommand{\ourshort}{ViSTA\xspace}
\newcommand{\ours}{ViSTA-SLAM\xspace}

\newcommand{\boldparagraph}[1]{\vspace{2pt}\noindent{\bf #1 \;\;}}

\definecolor{myPink}{rgb}{0.980, 0.502, 0.447} 
\definecolor{myOrange}{rgb}{1.000, 0.784, 0.486} 

\colorlet{colorFst}{myPink!40} 
\colorlet{colorSnd}{myOrange!45} 
\colorlet{colorTrd}{Goldenrod!20} 

\colorlet{colorLow}{darkgray!60}    
\newcommand{\st}{\cellcolor{colorFst}\bf}   
\newcommand{\nd}{\cellcolor{colorSnd}}      
\newcommand{\rd}{\cellcolor{colorTrd}}      
\newcommand{\lo}{\color{colorLow}}          

\definecolor{gray}{rgb}{0.65,0.65,0.65}
\definecolor{mycol}{rgb}{0.90,0.95,1.0}

\title{ViSTA-SLAM: Visual SLAM with Symmetric Two-view Association}

\author{Ganlin Zhang$^{~1,2}$
\quad
Shenhan Qian$^{~1,2}$
\quad
Xi Wang$^{~1,2,3}$
\quad
Daniel Cremers$^{~1,2}$
\vspace{0.3em}\\  $^{1~}$TU Munich \qquad $^{2~}$MCML \qquad $^{3~}$ETH Zurich
}
\begin{document}
\maketitle
\begin{strip}
  \vspace{-4em}
  \centering
  \footnotesize
  \captionsetup{type=figure} 
  \resizebox{\linewidth}{!}{
  \renewcommand{\arraystretch}{1}
  \includegraphics{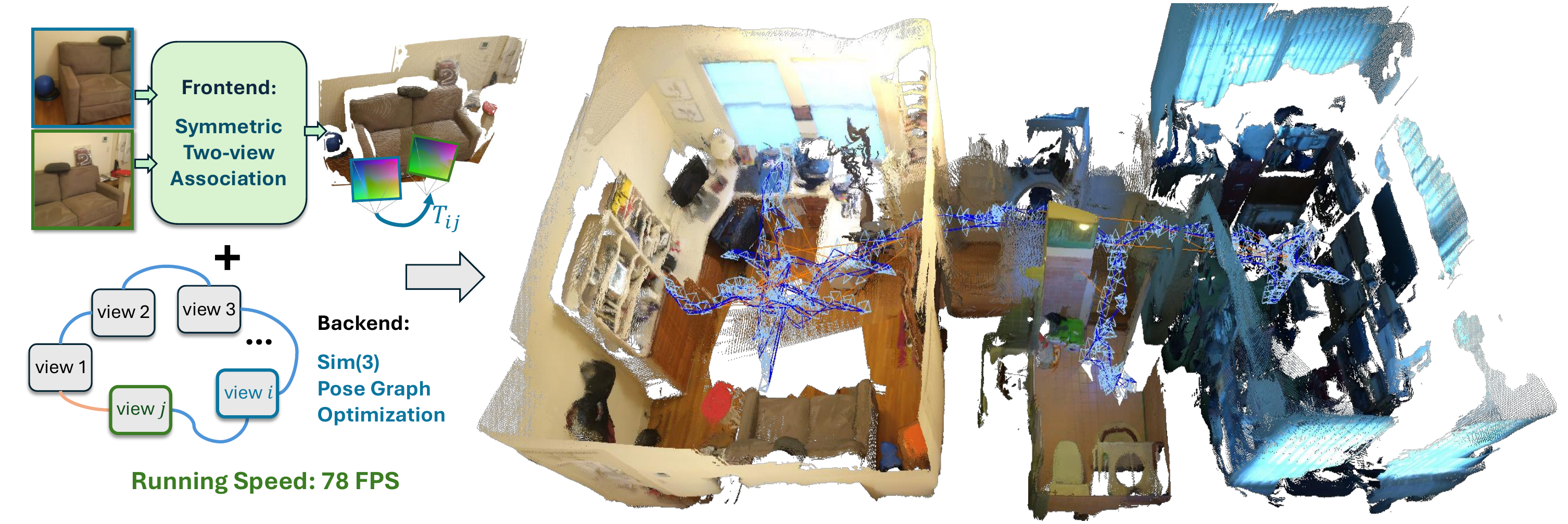}
  
  }
\captionof{figure}{\textbf{\ours Results on a Multi-room Scene~\cite{dai2017scannet}.}  
By combining the proposed lightweight frontend Symmetric Two-view Association (STA) model with $\mathrm{Sim(3)}$ pose graph optimization and loop closuring as the backend, \ours achieves high-quality reconstruction and accurate trajectory estimation on challenging scenes while running in real time. }
\label{fig:teaser}
\vspace{-0.2em}
\end{strip}

\begin{abstract}
\vspace{-0.7em}
We present \ours as a real-time monocular visual SLAM system that operates without requiring camera intrinsics, making it broadly applicable across diverse camera setups. At its core, the system employs a lightweight symmetric two-view association (STA) model as the frontend, which simultaneously estimates relative camera poses and regresses local pointmaps from only two RGB images. This design reduces model complexity significantly, the size of our frontend is only 35\% that of comparable state-of-the-art methods,
while enhancing the quality of two-view constraints used in the pipeline.
In the backend, we construct a specially designed $\mathrm{Sim}(3)$ pose graph that incorporates loop closures to address accumulated drift. Extensive experiments demonstrate that our approach achieves superior performance in both camera tracking and dense 3D reconstruction quality compared to current methods. Github repository: \url{https://github.com/zhangganlin/vista-slam}

\end{abstract}
\vspace{-1em}    
\vspace{-0.8em}
\section{Introduction}
\label{sec:intro}

\subsection{Real-time Monocular Dense SLAM}
Simultaneous Localization and Mapping (SLAM) jointly estimates an agent's pose and the surrounding 3D scene from sensor observations. In the monocular dense setting, a single RGB camera is used to reconstruct a continuous 3D map, enabling both geometric accuracy and visual realism. The camera poses and dense reconstruction outputs underpin downstream tasks~\cite{sengupta2013urban,armeni2019scenegraph,miao2024scenegraphloc,miao2024volumetric,halacheva2024holistic} such as semantic perception, object interaction, and scene editing, and are crucial for applications in VR/AR, robotics, and autonomous driving, where accurate, low-latency 3D perception is essential.

\subsection{Related Work}
\label{sec:related_work}
\boldparagraph{Classical Visual SLAM}
Classical visual SLAM methods can be broadly categorized into two types. The first, similar to incremental SfM~\cite{agarwal2011building,schonberger2016structure,liu2024robust}, is feature-based SLAM~\cite{davison2007monoslam,mur2015orb,mur2017orb,campos2021orb}, relying on keypoint extraction and descriptor matching to provide constraints for triangulation and PnP~\cite{lepetit2009ep,kneip2011novel} pose estimation. The second, known as direct methods, such as LSD-SLAM~\cite{engel2014lsd} and DSO~\cite{engel2017direct}, optimizes camera poses by minimizing photometric error from pixel intensities while estimating a per-frame depth map. Both categories typically adopt a frontend (feature-based or direct) and a backend for optimization, most often bundle adjustment~\cite{triggs1999bundle} to jointly refine poses and structure. However, they rely heavily on accurate camera calibration and are generally limited to sparse 3D reconstructions.

\boldparagraph{Dense Visual SLAM}
To enable denser reconstructions, recent works have incorporated deep learning into either the frontend or the scene representation. For example, DROID-SLAM~\cite{teed2021droid} uses RAFT~\cite{teed2020raft} for dense optical flow in the frontend and performs dense bundle adjustment on the GPU, while methods such as SuperPrimitive~\cite{mazur2024superprimitive} and COMO~\cite{dexheimer2024compact} leverage monocular priors (e.g., surface normals, depth distributions) from pretrained models with direct photometric optimization. BA-Track~\cite{chen2025back} augments point-based tracking with a scale-grid deformation of monocular depth priors. Neural scene representations have also been adopted, with NICER-SLAM~\cite{zhu2024nicer} and MonoGS~\cite{matsuki2024gaussian} optimizing camera poses and 3D structure using NeRF~\cite{mildenhall2021nerf} and 3D Gaussian Splatting~\cite{kerbl20233dgs}, respectively. Tracking robustness and geometric accuracy have been further improved by integrating depth priors and auxiliary trackers~\cite{zhang2024glorie,sandstrom2025splat,zhang2023go,zhang2023hi,zhang2024hi}. However, most approaches still require accurate camera intrinsics and many struggle to achieve true real-time performance due to the computational demands of dense optimization and neural rendering.

\boldparagraph{SLAM with 3D Foundation Models}
All aforementioned methods require known and accurate camera intrinsics. With the advent of 3D foundation models~\cite{dust3r,mast3r,wang2025vggt}, several intrinsic-free SLAM frameworks have emerged, aiming to produce dense outputs without calibration. Methods such as Spann3R~\cite{wang2025spann3r} and others~\cite{liu2025slam3r,wang2025cut3r,cabon2025must3r} extend the two-view DUSt3R~\cite{dust3r} model to sequential inputs, directly regressing point clouds in a unified global coordinate system. Reloc3r~\cite{dong2025reloc3r} instead regresses only relative poses and performs offline global optimization using SfM techniques~\cite{pan2024glomap,sweeney2015theia,zhang2023revisiting}. MASt3R-SLAM~\cite{murai2025mast3rslam} extracts dense correspondences from MASt3R~\cite{mast3r} and feeds them into a classical optimization pipeline, while submap-based methods~\cite{maggio2025vggtslam,deng2025vggtlong} employs the multiview model VGGT~\cite{wang2025vggt} to regress local submaps before stitching them via pose graph optimization. 
While these approaches address some classical limitations, they still face notable drawbacks:
\begin{enumerate}[label=\arabic*., leftmargin=*, itemsep=0em, parsep=0pt, topsep=0pt]
    \item Current two-view models~\cite{dust3r,mast3r} use asymmetric architectures that regress pointmaps of both views to the first view's coordinates, making it difficult to decouple views for backend optimization (\eg loop closure). 
    \item Pure regression methods~\cite{wang2025spann3r,wang2025cut3r,liu2025slam3r} predict incoming frames with previous memory, but suffer from drift and start forgetting once the trajectory gets longer. 
    \item Methods~\cite{wang2025spann3r,wang2025cut3r,liu2025slam3r,murai2025mast3rslam} built on current two-view models inherit the asymmetric architecture with two separate decoders, resulting in large model size. Submap-based methods~\cite{maggio2025vggtslam,deng2025vggtlong} employ an even larger multiview model~\cite{wang2025vggt} to build submaps, which further increases the size of the frontend model.
\end{enumerate}

\begin{figure*}[!t]
\centering
\includegraphics[width=1.0\linewidth]{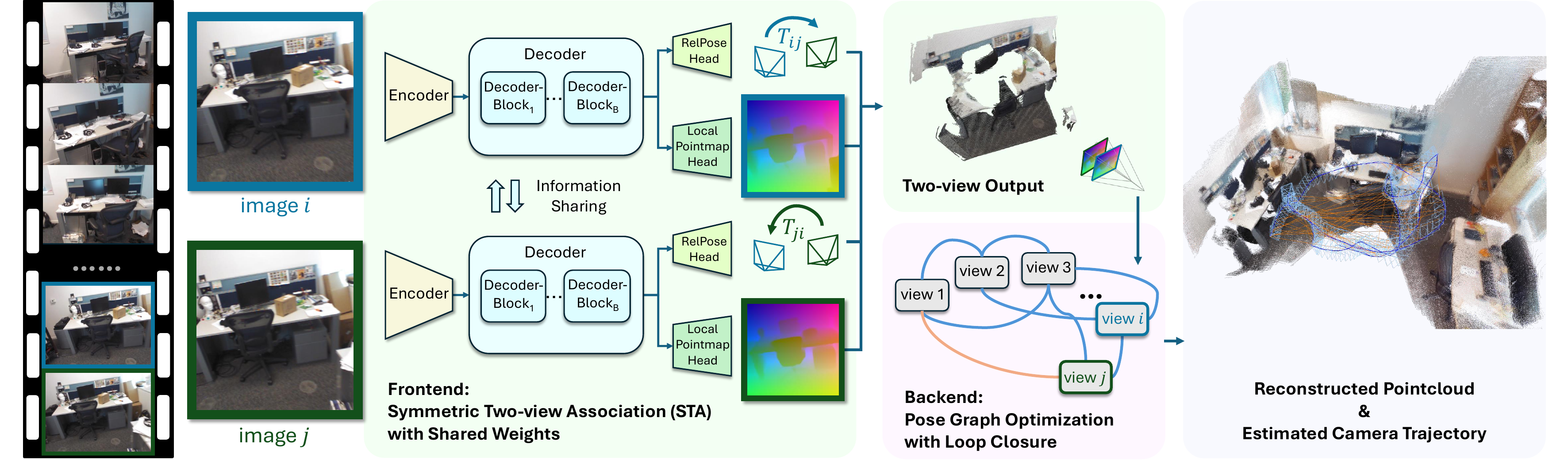}
\caption{\textbf{\ours Overview.} Given sequential video frames without intrinsics as the input, our frontend model takes in view pairs and predicts local pointmaps and relative poses within each pair. 
We then use the pair-wise predictions to construct a $\mathrm{Sim(3)}$ pose graph with loop closure and optimize it via Levenberg–Marquardt algorithm. The frontend model employs a fully symmetric design, making the model lightweight and supporting more flexible pose graph optimization. The blue edges in the pose graph and final results represent connections between neighboring nodes (views), while the orange edges correspond to loop closures.}
\label{fig:architecture}
\vspace{-0.5em}
\end{figure*}

\subsection{Contributions of \ours}

To address these concerns, we propose \ours, a novel real-time monocular visual SLAM pipeline based on symmetric two-view association. At its core is a lightweight Symmetric Two-view Association (STA) model frontend, which takes two RGB images as input and simultaneously regresses two pointmaps in their respective local coordinate frames, along with the relative camera pose between them. During training, we enforce cycle consistency on relative poses and geometric consistency on pointmaps to improve accuracy and stability. Unlike prior 3D models~\cite{dust3r,mast3r,wang2025vggt}, STA is \textit{fully symmetric} with respect to its inputs: neither view is designated as a reference, and the same encoder–decoder architecture is applied to both. In the backend, we perform $\mathrm{Sim}(3)$ pose graph optimization with loop closures to mitigate drift and ensure global consistency. To further enhance robustness, each view is represented by multiple nodes rather than a single one, which are connected by scale-only edges to handle scale inconsistencies across different forward passes.

This symmetric design makes our frontend substantially more lightweight than existing methods, with STA being only 64\% the size of MASt3R~\cite{mast3r} and 35\% the size of VGGT~\cite{wang2025vggt}. 
Unlike prior approaches~\cite{maggio2025vggtslam,deng2025vggtlong} that group multiple views into a single submap node, our method assigns each view its own nodes in the pose graph. Leveraging the local pointmaps produced by the STA frontend, each node can be represented independently, connected to others solely through relative transformations. Compared to submap-based methods~\cite{maggio2025vggtslam}, this design yields a more flexible graph structure and greater robustness. The combination of this flexibility and lightweight architecture underpins our choice of a symmetric two-view model as the frontend.

In summary, our main \textbf{contributions} are as follows:
\begin{itemize}[itemsep=0pt,topsep=2pt,leftmargin=10pt,label=$\bullet$] 
  \item We design and train a lightweight, symmetric two-view association network as the frontend, which takes only two RGB images as input and regresses their pointmaps in local coordinates along with the relative camera pose.
  \item We construct a robust $\mathrm{Sim}(3)$ pose graph with loop closures, optimize it using the Levenberg–Marquardt algorithm for fast and stable convergence.
  \item By integrating these components, we present a real-time monocular dense visual SLAM framework that operates without requiring any camera intrinsics.
  \item Our method achieves state-of-the-art performance on the real-world 7-Scenes~\cite{shotton2013-7scenes} and TUM-RGBD~\cite{sturm12tumrgbd} datasets for both camera trajectory estimation and dense 3D reconstruction.
\end{itemize}

\section{\ours Pipeline}
\label{sec:method}
As a monocular dense SLAM pipeline (\cref{fig:architecture}), our aim is to simultaneously track camera poses and reconstruct the recorded scene online using a dense pointcloud. To achieve this, we propose a lightweight and novel symmetric two-view association model as the frontend of our pipeline, which extracts the relative pose and local point maps of two neighboring input frames (\cref{subsec:frontend} and \cref{subsec:loss}). In the backend, a sparse and efficient $\mathrm{Sim}(3)$ pose graph optimization with loop closure is performed to mitigate drift accumulation (\cref{subsec:backend}).

\subsection{Symmetric Two-view Association Model}\label{subsec:frontend}
In classical monocular SLAM pipelines, the two-view estimation is one of the most critical building block, as it establishes geometric constraints that allow for further optimization. In this work, we follow the same principle; however, instead of relying on traditional methods, we propose a deep learning based Symmetric Two-view Association (STA) model that eliminates the need for camera intrinsics in the SLAM process. 

\boldparagraph{Encoder}
Our STA model takes in two images $\bm{I}_i$, $\bm{I}_j$ as input. It uses a shared ViT encoder~\cite{dosovitskiy2020image} to divide input images into patches and encode them into features
\begin{equation*}
    \bm E_{i/j} = \text{Encoder}\left(\bm I_{i/j}\right) \quad \in \mathbb{R}^{K\times C},
\end{equation*}
where $K$ represents tokens and $C$ denotes the dimensions. Subsequently, we insert a camera pose embedding $\bm p$ to the encoding features of each view, forming
\begin{equation*}
    \bm E'_{i/j} = \left(\bm p, \bm E_{i/j}\right) \quad \in \mathbb{R}^{(K+1)\times C}.
\end{equation*}

\boldparagraph{Decoder}
The decoder further processes and fuses information between the encoding features $\bm E'_i$ and $\bm E'_j$. It contains a sequence of $B$ decoder blocks, each conducting a self-attention operation followed by a cross-attention operation, producing decoding features
\begin{align*}
\bm D_{i/j}^{(b)} &= \text{DecoderBlock}^{(b)} \left(\bm D_{i/j}^{(b-1)}, \bm D_{j/i}^{(b-1)}\right) \in \mathbb{R}^{(K+1) \times C'},
\end{align*}
where $b \in \left\{1, 2, \dots, B\right\}$ is the index of a decoder block, and $\bm D_{i/j}^{(0)}=\bm E'_{i/j}$.

\begin{figure}[!t]
\centering

\includegraphics[width=\linewidth]{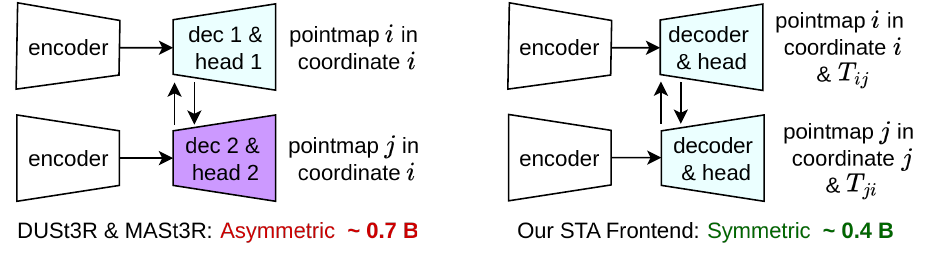}
\caption{\textbf{Asymmetric vs. Symmetric Architectures.} 
Asymmetric architectures~\cite{dust3r,mast3r} use two decoders to regress point maps in a shared coordinate space. our symmetric formulation regresses relative pose and local point maps with only a single decoder, reducing over 36\% of the parameters ({$\sim$} 0.4 vs. 0.7 billion),  while achieving higher accuracy and enabling pose graph optimization in the backend.}
\label{fig:symmetric}
\vspace{0.5em}
\includegraphics[width=0.9\linewidth]{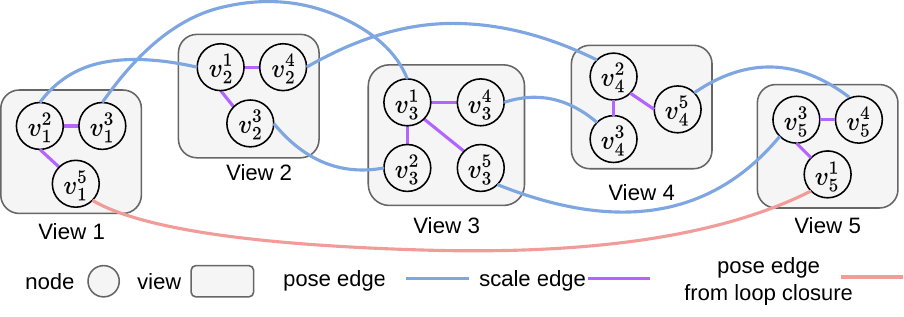}
\caption{\textbf{An Example Pose Graph for 5 views $(N=2)$.} Nodes of the same view are grouped and connected to the first processed node of that view via \textit{scale edges}. Our two-type-edge design enhances optimization robustness, yielding more accurate poses.}
\label{fig:posegraph}
\vspace{-0.5em}
\end{figure}

\boldparagraph{Symmetric Formulation}
Prior approaches~\cite{dust3r,mast3r} regress both point maps into the coordinate frame of the first view, thus requiring two separate decoders. In contrast, our model predicts only local point maps and relative poses between views. This fully symmetric formulation makes it possible to use only one decoder. As a result, the number of parameters for decoding is effectively reduced by half (shown in \cref{fig:symmetric}), forming a more compact model for real-time applications. Moreover, producing local view outputs in their own coordinate systems is better suited for the subsequent pose graph optimization, see \cref{subsec:backend} for details.

\boldparagraph{Local Point Maps}
Given decoding features $\bm D_{i/j}^{(b)}$, we use a DPT head~\cite{ranftl2021vision} to regress the local point maps $\bm P$ and corresponding confidence maps $\bm W$:
\begin{align*}
    \bm P_{i/j}, \bm W_{i/j} = \text{PointHead}\left(\bm D_{i/j}^{(b)}\right)
\end{align*}

\boldparagraph{Relative Poses}
Given the first embedding of the decoding feature $\bm D_i^{(B)}$, i.e., the camera pose embedding $\bm{p}_i^{(B)} \in \mathbb{R}^{1\times C'}$, we use an MLP to regress the relative transformation from view $i$ to $j$ . Specifically, the MLP outputs a matrix $\bm M_{ij} \in \mathbb{R}^{3\times 3}$ for rotation, a translation vector $\bm t_{ij} \in \mathbb{R}^{3\times1}$, and a confidence score $w_{ij} \in [0,1]$:
\begin{equation*}
    \bm M_{ij}, \bm t_{ij}, w_{ij} = \text{PoseHead}(\bm p_i^{(B)})
\end{equation*}
Since $\bm M_{ij}$ is not guaranteed to lie on the $SO(3)$ manifold, we apply SVD orthogonalization~\cite{levinson2020analysis} to it to obtain a valid rotation matrix $\bm R_{ij}$. 
Then, the relative transformation is $\bm T_{ij} = \left[\bm R_{ij} | \bm t_{ij}\right]$. 
With our symmetric formulation, we could also input $\bm p_j$ into the pose head to regress $\bm T_{ji}$. But in practice we only regress one of them for building pose graph.

\subsection{Training Objective}
\label{subsec:loss}
There are three loss terms to supervise the training of our STA model. \textit{Pointmap Loss} compares the regressed local pointmap with ground-truth points. \textit{Relative Pose Loss} penalizes errors in relative rotation and translation, with a cycle-consistency term ensuring the two predicted poses are mutual inverses. \textit{Geometric Consistency Loss} enforces alignment of the two local point maps after applying the predicted relative transformation, improving local reconstruction consistency.

\boldparagraph{Local Pointmap Loss}
Following DUSt3R~\cite{dust3r}, we apply the confidence-weighted regression loss for all predicted point maps with valid ground truths. Since the reconstruction is up-to-scale, we also normalize the regressed pointmap and the ground-truth pointmap according to their mean Euclidean distance to the origin, $n$ and $\hat{n}$:
\begin{align}
L_\text{pmap} = \sum_{v \in \{i, j\}} \sum_{\bm x \in \bm{I}_v} & \left\|\bm W_v(\bm x)\cdot \left(\frac{\hat{\bm{P}}_v(\bm x)}{\hat{n}} - \frac{\bm{P}_v(\bm x)}{n} \right)\right\|   \nonumber\\
&- \alpha^{\text{point}}\log\left(\bm W_v(\bm x)\right),
\label{eq:loss_pmap}
\end{align}
where $x$ is the pixel coordinate. Note that all the points are regressed in the local coordinate space of each view. 

\boldparagraph{Relative Pose Loss}
Relative pose loss consists of three parts: \textit{rotation loss}, \textit{translation loss} and \textit{identity loss}.
The rotation loss $L_R$ evaluates the angle between the regressed rotation $\bm R_{ij}$ and the ground-truth rotation $\hat{\bm R}_{ij}$,
\begin{equation}
    L_R(\bm R, \hat{\bm R}) = \arccos\left(\dfrac{\text{tr}(\bm R^{-1}\hat{\bm R})-1}{2}\right).
\end{equation}
The translation loss $L_t$ evaluates the euclidean distance between the predicted translation $\bm t_{ij}$ and the ground truth $\hat{\bm t}_{ij}$, which are normalized by the same factors $n$ and $\hat{n}$ as for the pointmap loss in \cref{eq:loss_pmap}:
\begin{equation}
    L_t(\bm t,\hat {\bm t}) = \left\| \frac{\bm t}{n} - \frac{\hat{\bm t}}{\hat{n}} \right\|^2.
\end{equation}
The pose identity loss $L_{id}$ minimizes the difference of $\bm T_{ij} \bm T_{ji}$ and the identity transformation $\bm I$, essentially constraining $\bm T_{ij}$ and $\bm T_{ji}$ to be the inverse of each other to improves the consistency of our pose prediction:
\begin{equation}
    L_{id} = L_R(\bm R_{ij}\bm R_{ji}, \bm I) + L_t(\bm R_{ij}\bm t_{ji}+\bm t_{ij}, \bm 0).
\end{equation}
Then the complete relative pose loss is defined as
\begin{align}
    L_\text{pose} = &w_{ij}\left(L_R(\bm R_{ij},\hat{\bm R}_{ij}) + L_t(\bm t_{ij},\hat {\bm t}_{ij})+L_{id}\right) \nonumber\\
    &- \alpha\log(w_{ij}), 
\end{align}
weighted by a separate confidence score $w_{ij}$ for pose regression.

\boldparagraph{Geometrical Consistency Loss}
To ensure that the predicted point maps of each pair are spatially consistent when placed in the same coordinate space, we introduce a \textit{geometric consistency loss}.
Given a pair of images with ground-truth intrinsics, depths, and relative poses, each pixel $\bm x$ in view $i$ can be accurately projected into view $j$, yielding its ground-truth corresponding pixel $\bm C_{ij}(\bm{x})$ in view $j$. Then the geometric consistency loss $L_\text{gc}$ is defined as:
\begin{equation}
    L_\text{gc} = \sum_{\bm x \in \bm{I}_{i}} \left\| \bm T_{ij}  \bm P_i(\bm x) - \bm P_j\left( \bm C_{ij}(\bm x) \right) \right\|/n,
\end{equation}
where $\bm T_{ij}$ is the predicted relative transformation and $n$ is the same normalization factor as in \cref{eq:loss_pmap}.

The total training objective function $L$ is the weighted sum of the above three losses,
\begin{equation}
    L = \lambda_\text{pmap}L_\text{pmap} + \lambda_\text{pose}L_\text{pose} + \lambda_\text{gc}L_\text{gc},
\end{equation}
where $\lambda_\text{pmap}$, $\lambda_\text{pose}$ and $\lambda_\text{gc}$ are weighting factors.

\begin{table*}[t]
    \centering
    \vspace{-0.5em}
    \small
    \setlength{\tabcolsep}{8.7pt}
    {
    \begin{tabular}{ll|ccccccc|c}
    \toprule
    & Method  & \texttt{chess} & \texttt{fire} & \texttt{heads} & \texttt{office}& \texttt{pumpkin} & \texttt{kitchen} & \texttt{stairs} & \textbf{Avg.}\\
    \midrule
    \cellcolor[HTML]{EEEEEE} & \lo NICER-SLAM~\cite{zhu2024nicer}
    & \lo \textbf{0.033} & \lo 0.069 & \lo 0.042 & \lo 0.108 & \lo 0.200 & \lo 0.039 & \lo 0.108 & \lo 0.086\\
    \cellcolor[HTML]{EEEEEE} & \lo DROID-SLAM~\cite{teed2021droid} 
    & \lo 0.036& \lo \textbf{0.027} & \lo 0.025 & \lo \textbf{0.066} & \lo \textbf{0.127} & \lo \textbf{0.040} & \lo 0.026 & \lo \textbf{0.049}\\
    \multirow{-3}{*}{\cellcolor[HTML]{EEEEEE}\rotatebox[origin=c]{90}{\textit{Calib.}}} & \lo GlORIE-SLAM~\cite{zhang2024glorie}
    & \lo 0.036 & \lo 0.029 & \lo \textbf{0.014} & \lo 0.094 & \lo 0.144 & \lo 0.053 & \lo \textbf{0.020} & \lo 0.056\\
    \midrule
    \cellcolor[HTML]{EEEEEE} & CUT3R~\cite{wang2025cut3r}  
    & 0.743 & 0.226 & 0.363 & 0.664 & 0.546 & 0.381 & 0.413 & 0.477\\
    \cellcolor[HTML]{EEEEEE} & SLAM3R~\cite{liu2025slam3r}   
    & 0.089 & 0.048 & 0.036 & \nd 0.088 & 0.196 & 0.102 & 0.126 & 0.098\\
    \cellcolor[HTML]{EEEEEE} & MASt3R-SLAM~\cite{murai2025mast3rslam}   
    & \nd 0.063 & \rd 0.046 & \rd 0.029 & \rd0.103 & \st 0.112 & \rd 0.074 & \nd 0.032 & \nd 0.066\\
    \cellcolor[HTML]{EEEEEE} & VGGT-SLAM~\cite{maggio2025vggtslam}   
    & \st 0.037 & \st 0.026 & \st 0.022 & \rd 0.103 & \rd 0.147 & \nd 0.063 & \rd 0.095 & \rd 0.070\\
    \multirow{-5}{*}{\cellcolor[HTML]{EEEEEE}\rotatebox[origin=c]{90}{\textit{Uncalib.}}} & \textbf{\ours} 
    & \rd 0.073 & \nd 0.035 & \nd 0.028 & \st 0.055 & \nd 0.129 & \st 0.035 & \st 0.029 & \st 0.055\\
    \bottomrule
    \end{tabular}
    }
    \caption{
    \textbf{Camera Trajectory Estimation (ATE RMSE) on 7-Scenes~\cite{shotton2013-7scenes}.}
    \ours performs the best on average.
    }
    \label{tab:traj_7scenes}
    \vspace{-1em}
\end{table*}

\subsection{Backend Pose Graph Optimization}
\label{subsec:backend}
\boldparagraph{Notation} In the backend, a $\mathrm{Sim}(3)$ pose graph $\mathcal{G} = (\mathcal{V}, \mathcal{E})$ is maintained and optimized to mitigate accumulated errors introduced by the two-view estimations. The vertex set $\mathcal{V}$ and edge set $\mathcal{E}$ are defined as
\begin{align}
    \mathcal{V} = \{\bm v_i^j\mid i,j \in\mathbb{N} \}, \quad
    \mathcal{E} = \{\bm e_{ij} \mid i,j\in\mathbb{N} \},
\end{align}
 where $\bm v_i^j,\bm e_{ij} \in \mathrm{Sim}(3)$, each vertex $\bm v_i^j$ represents an absolute camera pose with scale, of view $i$, with view $i$ and view$j$ as the input of STA; each edge $\bm e_{ij}$ encodes the relative transformation between a pair of connected vertices $\bm v_i^j$ and $\bm v_j^i$. 
 Both $\bm v_i^j$ and $\bm e_{ij}$ have a rigid transformation part $\bm T\in \mathrm{SE}(3)$  and a scale part $s\in\mathbb{R}^+$. 

\boldparagraph{Graph Construction}
To construct the pose graph, for a given view $i$, the STA model performs $2N$ forward passes with neighboring views $j \in [i-N, i-1] \cup [i+1, i+N]$. The predicted pointmap for each view in each forward pass corresponds to a node in the graph, resulting in multiple nodes per view since each view is processed multiple times by the frontend model.
As shown in \cref{fig:posegraph}, we define two types of edges in the graph. \textit{Pose edges} connect two nodes generated from the same forward pass, using the estimated relative camera pose and an identity relative scale, based on the assumption that the two local point maps from a single forward pass share the same scale. \textit{Scale edges} connect nodes belonging to the same view but obtained in different forward passes (paired with different neighboring views), with the rigid transformation component set to identity, and the scale component solved via weighted least squares between the predicted point maps of different forward passes. 
Among all nodes of the same view, \textit{scale edges} are constructed only between the first processed node and the others for sparsity and simplicity.

\boldparagraph{Loop Closure} We use Bag of Words~\cite{GalvezTRO12} to detect loop candidates, which form new pairs. Then we can feed each candidates pair into our STA model to confirm these proximity. If the predicted confidence score of the relative pose is higher than a predefined threshold $\tau_p$, this pair is accepted as a valid loop, and two new nodes connected by a \textit{pose edge} are added. Each new node is also connected to the first processed node of its corresponding view via a \textit{scale edge}.

\boldparagraph{Optimization}
We perform pose graph optimization using the Levenberg--Marquardt algorithm in the space of Lie algebra $\mathfrak{sim}(3)$.
\begin{equation}
\min_{\{\bm v_i^j \in \mathcal{V}\}} \sum_{\bm e_{ij} \in \mathcal{E}} 
\left\|
\log_{\mathrm{Sim}(3)}\left( \bm e_{ij} \cdot ({\bm{v}}_i^j)^{-1}\cdot {\bm{v}}_j^i \right)
\right\|_{\bm \Omega_{ij}}^2
\end{equation}
where $\bm \Omega_{ij}$ represents the covariance matrix, derived from the confidence score predicted by the STA model. The optimization process takes less than 5 iterations to converge in most cases.

Using the optimized camera pose and scale, the reconstructed pointcloud $\tilde{\bm P}_i^j$ in global coordinate is,
\begin{equation}
    \tilde{\bm P}_i^j = s_i^j \bm R_i^j \bm P_i^j + \bm t_i^j,
\end{equation}
where $\bm R_i^j$, $\bm t_i^j$ and $s_i^j$ are the orientation, position, scale of $\bm v_i^j$ respectively. To avoid redundancy, for each view $i$, we only keep the pointcloud with largest confidence among all $\tilde{\bm P}_i^*$ in the final result.

\begin{table*}[t]
    \centering
    \small
    \setlength{\tabcolsep}{7.5pt}
    \begin{tabular}{ll|ccccccccc|c}
        \toprule
        & Method  & \texttt{360} & \texttt{desk} & \texttt{desk2} & \texttt{floor}& \texttt{plant} & \texttt{room} & \texttt{rpy} & \texttt{teddy} & \texttt{xyz} & \textbf{Avg.} \\
        \midrule
        
        \cellcolor[HTML]{EEEEEE} & \lo ORB-SLAM3~\cite{campos2021orb}
        & \lo $\times$ & \lo 0.017 & \lo 0.210 & \lo $\times$ & \lo 0.034 & \lo $\times$ & \lo $\times$ & \lo $\times$ & \lo \textbf{0.009} & \lo {N/A} \\
        \cellcolor[HTML]{EEEEEE} & \lo DPV-SLAM~\cite{lipson2024dpvslam} 
        & \lo 0.112 & \lo 0.018 & \lo 0.029 & \lo 0.057 & \lo 0.021 & \lo 0.330 & \lo 0.030 & \lo 0.084 & \lo 0.010 & \lo 0.076 \\
        \cellcolor[HTML]{EEEEEE} & \lo DPV-SLAM++~\cite{lipson2024dpvslam} 
        & \lo 0.132 & \lo 0.018 & \lo 0.029 & \lo 0.050 & \lo 0.022 & \lo 0.096 & \lo 0.032 & \lo 0.098 & \lo 0.010 & \lo 0.054 \\
        \cellcolor[HTML]{EEEEEE} & \lo DROID-SLAM~\cite{teed2021droid} 
        & \lo \textbf{0.111} & \lo 0.018 & \lo 0.042 & \lo \textbf{0.021} & \lo \textbf{0.016} & \lo 0.049 & \lo 0.026 & \lo 0.048 & \lo 0.012 & \lo 0.038 \\
        \multirow{-5}{*}{\cellcolor[HTML]{EEEEEE}\rotatebox[origin=c]{90}{\textit{Calibrated}}}  & \lo GlORIE-SLAM~\cite{zhang2024glorie}
        & \lo 0.128 & \lo \textbf{0.016} & \lo \textbf{0.028} & \lo \textbf{0.021} & \lo 0.021 & \lo \textbf{0.042} & \lo \textbf{0.020} & \lo \textbf{0.035} & \lo 0.010 & \lo \textbf{0.036} \\
        \midrule
        
        \cellcolor[HTML]{EEEEEE} & CUT3R~\cite{wang2025cut3r}  
        & 0.174 & 0.592 & 0.546 & 0.662 & 0.467 & 0.911 & 0.051 & 0.845 & \rd 0.129 & 0.486  \\
        \cellcolor[HTML]{EEEEEE} & SLAM3R~\cite{liu2025slam3r}   
        & 0.211 & 0.861 & 0.967 & 0.790 & 0.755 & 1.013 & 0.063 & 0.986 & 0.185 & 0.648 \\
        \cellcolor[HTML]{EEEEEE} & MASt3R-SLAM~\cite{murai2025mast3rslam}   
        & \nd 0.070 & \rd 0.032 & \rd 0.055 & \st 0.056 & \nd 0.035 & \nd 0.118 & \rd 0.041 & \rd 0.116 & \nd 0.020 & \nd 0.060 \\
        \cellcolor[HTML]{EEEEEE} & VGGT-SLAM~\cite{maggio2025vggtslam}   
        & \st 0.063 & \nd 0.031 & \nd 0.048 & \rd 0.152 & \st 0.023 & \rd 0.133 & \nd 0.038 & \st 0.039 & \nd 0.020 & \rd 0.061 \\
        \multirow{-5}{*}{\cellcolor[HTML]{EEEEEE}\rotatebox[origin=c]{90}{\textit{Uncalibrated}}} & \textbf{\ours} 
        & \rd 0.104 & \st 0.030 & \st 0.030 & \nd 0.070 & \rd 0.052 & \st 0.067 & \st 0.023 & \nd 0.080 & \st 0.015 & \st 0.052 \\
        \bottomrule
    \end{tabular}
    \caption{
        \textbf{Camera Trajectory Estimation (ATE RMSE) on TUM-RGBD~\cite{sturm12tumrgbd}.}
        \ours performs the best on average.
    }
    \label{tab:traj_tumrgbd}
    \vspace{-0.5em}
\end{table*}

\vspace{-0.1em}
\section{Experiments}
\vspace{-0.1em}
\label{sec:exp}

\boldparagraph{Evaluation Datasets and Metrics}
Following VGGT-SLAM~\cite{maggio2025vggtslam}, we evaluate our method on standard monocular SLAM benchmarks for camera tracking accuracy and reconstruction quality. We report root mean square error (RMSE) of absolute trajectory error (ATE, in meters) on real-world 7-Scenes~\cite{shotton2013-7scenes} and TUM-RGBD~\cite{sturm12tumrgbd} datasets using the evo toolkit~\cite{grupp2017evo}. Reconstruction quality on 7-Scenes is assessed via RMSE of accuracy, completion, and Chamfer distance (meters), leveraging its ground-truth 3D scenes.

\boldparagraph{Implementation Details}
The frontend STA model is initialized from the weights of DUSt3R~\cite{dust3r}, and trained on  ScanNet~\cite{dai2017scannet}, ScanNet++\cite{yeshwanthliu2023scannetpp}, ARKitScenes\cite{dehghan2021arkitscenes}, CO3D~\cite{reizenstein21co3d}, Aria Synthetic Environments~\cite{avetisyan2024ase}, and Replica~\cite{replica19arxiv} for 7 days using 8 NVIDIA H100 GPUs.
We use AdamW optimizer to train our STA model with learning rate $1.5e^{-5}$, weight decay $0.01$, $B=12$, $\alpha^{\text{point}}=0.2$, $\alpha^{\text{pose}}=0.05$, $\lambda_{\text{pmap}}=1$, $\lambda_{\text{pose}}=1$, $\lambda_{\text{gc}}=1$, and $\tau_p = 0.75$. We conducted evaluations on a machine with an NVIDIA RTX 4090 GPU and an Intel i9-14900KF CPU, with $N=2$ for 7-Scenes and $N=3$ for TUM-RGBD.


\boldparagraph{Baselines}
\ours is primarily compared with state-of-the-art (SOTA) learning-based SLAM methods in uncalibrated scenarios: VGGT-SLAM~\cite{maggio2025vggtslam}, MASt3R-SLAM~\cite{murai2025mast3rslam}, SLAM3R~\cite{liu2025slam3r}, and CUT3R~\cite{wang2025cut3r}. To reduce randomness from VGGT-SLAM’s RANSAC, we run it 5 times per scene as suggested in their paper; results for other methods, including ours, are deterministic. We also compare with SOTA methods using known camera intrinsics~\cite{campos2021orb,zhu2024nicer,zhang2024glorie,teed2021droid,lipson2024dpvslam}. Some results are taken from~\cite{murai2025mast3rslam,maggio2025vggtslam}. For MASt3R-SLAM and VGGT-SLAM, we keep their original keyframe selection; for ours, CUT3R, and SLAM3R, we use frame strides of 5 (7-Scenes) and 3 (TUM-RGBD). Calibrated methods are shown in {\lo gray}. Best results are highlighted as \colorbox{colorFst}{\bf first}, \colorbox{colorSnd}{second}, and \colorbox{colorTrd}{third}.

\subsection{Camera Trajectory Evaluation}
\label{subsec:traj}

In \cref{tab:traj_7scenes} and \cref{tab:traj_tumrgbd}, we report ATE RMSE. \ours achieves the best average performance on both datasets, outperforming current SOTA~\cite{murai2025mast3rslam} by 17\% (0.055 vs. 0.066) and 13\% (0.052 vs. 0.060), and surpasses some calibrated methods~\cite{zhu2024nicer,lipson2024dpvslam}. \ours performs less effectively on TUM-RGBD  \texttt{360} scene due to predominantly rotational camera motion that leads to frontend ambiguity and degrading performance.
Other methods~\cite{murai2025mast3rslam,maggio2025vggtslam} use either heavier multi-view frontend or more intensive optimization to reduce the influence. 
Pure regression-based methods~\cite{wang2025cut3r,liu2025slam3r} struggle to maintain consistent registration over long sequences with large camera motion due to forgetting effects.

\begin{figure*}[!htb]
\centering
{\footnotesize
    \setlength{\tabcolsep}{0pt}
    \renewcommand{\arraystretch}{1}
    \newcommand{\sz}{1}
    \begin{tabular}{c}
      \vspace{-1em}\includegraphics[width=\sz\linewidth]{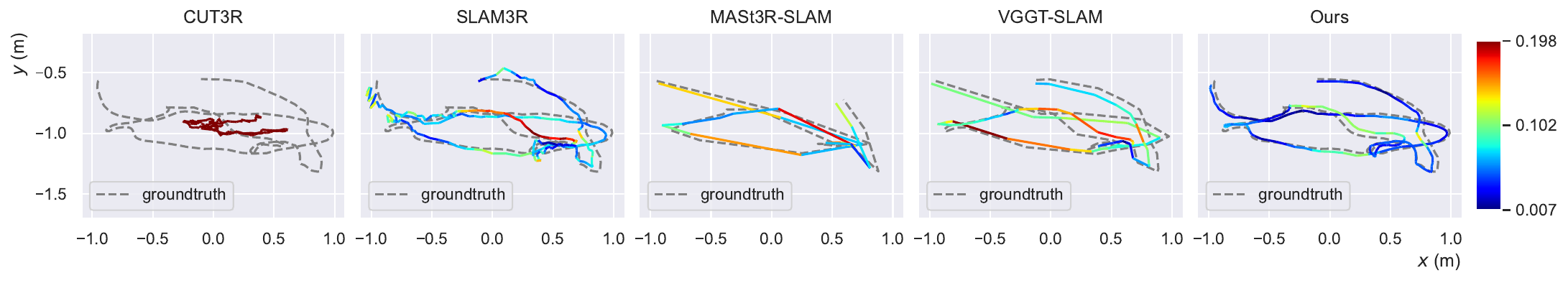} \\ [-12pt]
     \includegraphics[width=\sz\linewidth]{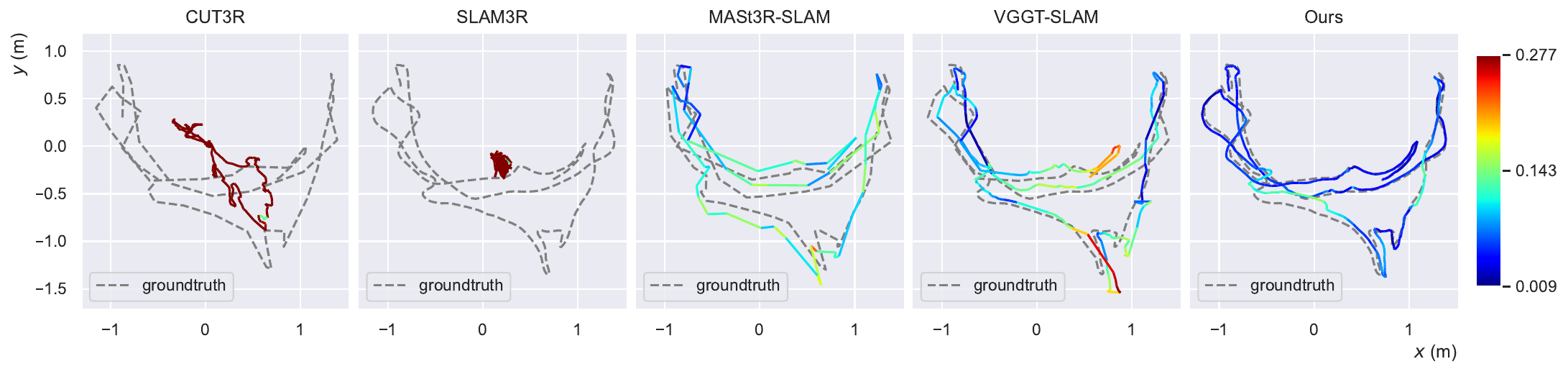} \\ [-8pt]

    \end{tabular}
}

\caption{\textbf{Trajectory estimation results on 7-Scenes \texttt{office} (top) and TUM-RGBD \texttt{room} (bottom).}
Estimated camera trajectories are projected onto the $x$–$y$ plane, with ground-truth shown as dashed lines. The trajectory color encodes ATE RMSE: higher errors in red, lower in blue. For MASt3R-SLAM~\cite{murai2025mast3rslam} and VGGT-SLAM~\cite{maggio2025vggtslam}, only the poses of their selected keyframes are estimated.}

\label{fig:traj}
\vspace{-1em}
\end{figure*}

In \cref{fig:traj}, we show the estimated trajectories from different methods on 7-Scenes~\cite{shotton2013-7scenes} \texttt{office} and TUM-RGBD~\cite{sturm12tumrgbd} \texttt{room}. CUT3R~\cite{wang2025cut3r} suffers from severe forgetting issues on long sequences; SLAM3R~\cite{liu2025slam3r} has bad point registration on the challenged scene TUM-RGBD \texttt{room}, thus, does not produce correct camera poses. Compared to pure regression-based methods, MASt3R-SLAM~\cite{murai2025mast3rslam} and VGGT-SLAM~\cite{maggio2025vggtslam} work well, while \ours achieves even higher trajectory accuracy.

\subsection{Dense Reconstruction Evaluation}

\label{subsec:recon}

\begin{table}[t]
    \centering
    \vspace{0.5em}
    \small
    \setlength{\tabcolsep}{4pt}
    \resizebox{\columnwidth}{!}
    {
    \begin{tabular}{l|c|ccc}
    \toprule
    Method  & ATE $\downarrow$ & Acc. $\downarrow$ & Comp. $\downarrow$ & Chamfer $\downarrow$\\
    \midrule
    \lo DROID-SLAM~\cite{teed2021droid} 
    &\lo 0.049 &\lo 0.141 &\lo\bf 0.048 &\lo 0.094\\
    \lo MASt3R-SLAM~\cite{murai2025mast3rslam} 
     &\lo\bf0.047 &\lo\bf 0.089 &\lo 0.085 &\lo\bf 0.087\\
    \midrule
    Spann3R @20~\cite{wang2025spann3r}
    & N/A & 0.069 & \nd 0.047 & 0.058\\
    Spann3R @2~\cite{wang2025spann3r}
    & N/A & 0.124 & \st 0.043 & 0.084\\
    CUT3R~\cite{wang2025cut3r}  
    & 0.477 & 0.276 & 0.303 & 0.290\\
    SLAM3R~\cite{liu2025slam3r}   
    & 0.098 & 0.053 & 0.059 & \nd 0.056\\
    MASt3R-SLAM~\cite{murai2025mast3rslam}   
    &\rd 0.066  & 0.059 & \rd 0.056 & \rd 0.057\\
    VGGT-SLAM~\cite{maggio2025vggtslam}   
    &  0.070  & \rd 0.052 & 0.060 & \nd 0.056\\
    2-view VGGT \emph{w/} PGO 
    &\nd 0.065 & \st 0.039 & 0.077 & 0.058\\
    \textbf{\ours} 
    & \st 0.055  & \nd 0.045 & \rd 0.056 & \st 0.051\\
    \bottomrule
    \end{tabular}
    }
    \caption{
    \textbf{Tracking and Reconstruction Evaluation on 7-Scenes~\cite{shotton2013-7scenes}.} @$n$ indicates a keyframe every $n$ images. \textit{2-view VGGT w/ PGO} uses the 2-view VGGT frontend with the same pose graph optimization as ours. \ours achieves the best trajectory estimation and reconstruction performance on 7-Scenes.
    }
    \label{tab:recon_7scenes}
    \vspace{2em}

    \centering
    \small
    \setlength{\tabcolsep}{3pt}
    \resizebox{\columnwidth}{!}
    {
    \begin{tabular}{lcccccc}
    \toprule
      & CUT3R & SLAM3R
      & MASt3R & VGGT &  2-view VGGT & \bf \ourshort\\
      & \cite{wang2025cut3r} & \cite{liu2025slam3r}& SLAM~\cite{murai2025mast3rslam} & SLAM~\cite{maggio2025vggtslam}& \emph{w/} PGO &\bf SLAM\\
    \midrule
    Size $\downarrow$ & 0.79 & \rd 0.76 & \nd 0.69 & 1.26 & 1.26 & \st0.44\\[2pt] \hdashline \noalign{\vskip 2pt}
    FPS $\uparrow$ & 34.2 & \rd 45.8 &30.3 & \st 93.3 & 12.6 & \nd 78.0\\
    \bottomrule
    \end{tabular}
    }
    \caption{
        \textbf{Model Size and Running Time Evaluation on 7-Scenes~\cite{shotton2013-7scenes} \texttt{redkitchen}.} Model sizes are reported in billions of parameters. FPS indicates the average number of frames processed per second over three runs. \ours shows highly competitive real-time speed with the smallest model size among baselines.
    }
    \label{tab:size_and_time}
    \vspace{-1em}
    
\end{table}


In \cref{tab:recon_7scenes}, we evaluate reconstruction quality across methods. Leveraging accurate camera poses and consistent local point clouds, \ours achieves the best Chamfer distance among all approaches. Despite using a lightweight two-view frontend, \ours, combined with tailored $\mathrm{Sim}(3)$ pose graph optimization, significantly outperforms multi-view-frontend methods~\cite{liu2025slam3r,maggio2025vggtslam} in accuracy (0.45 vs. 0.52) while matching or exceeding completeness. 
To demonstrate the effectiveness of our lightweight frontend, we add another strong baseline, replacing our STA model with a two-view VGGT~\cite{wang2025vggt} as the frontend and conducting the same pose graph optimization. \ours still achieves better performance in Chamfer distance, completeness, and absolute trajectory error, highlighting the effectiveness of our lightweight symmetric frontend over larger multiview models like VGGT for SLAM tasks

\begin{figure*}[!ht]
\centering
{\footnotesize
    \setlength{\tabcolsep}{3pt}
    \renewcommand{\arraystretch}{1}
    \newcommand{\sz}{0.3}
    \begin{tabular}{cccc}
    \rotatebox{90}{\hspace{1.2em} \small \makecell{CUT3R~\cite{wang2025cut3r}}} &
        \includegraphics[width=\sz\linewidth]{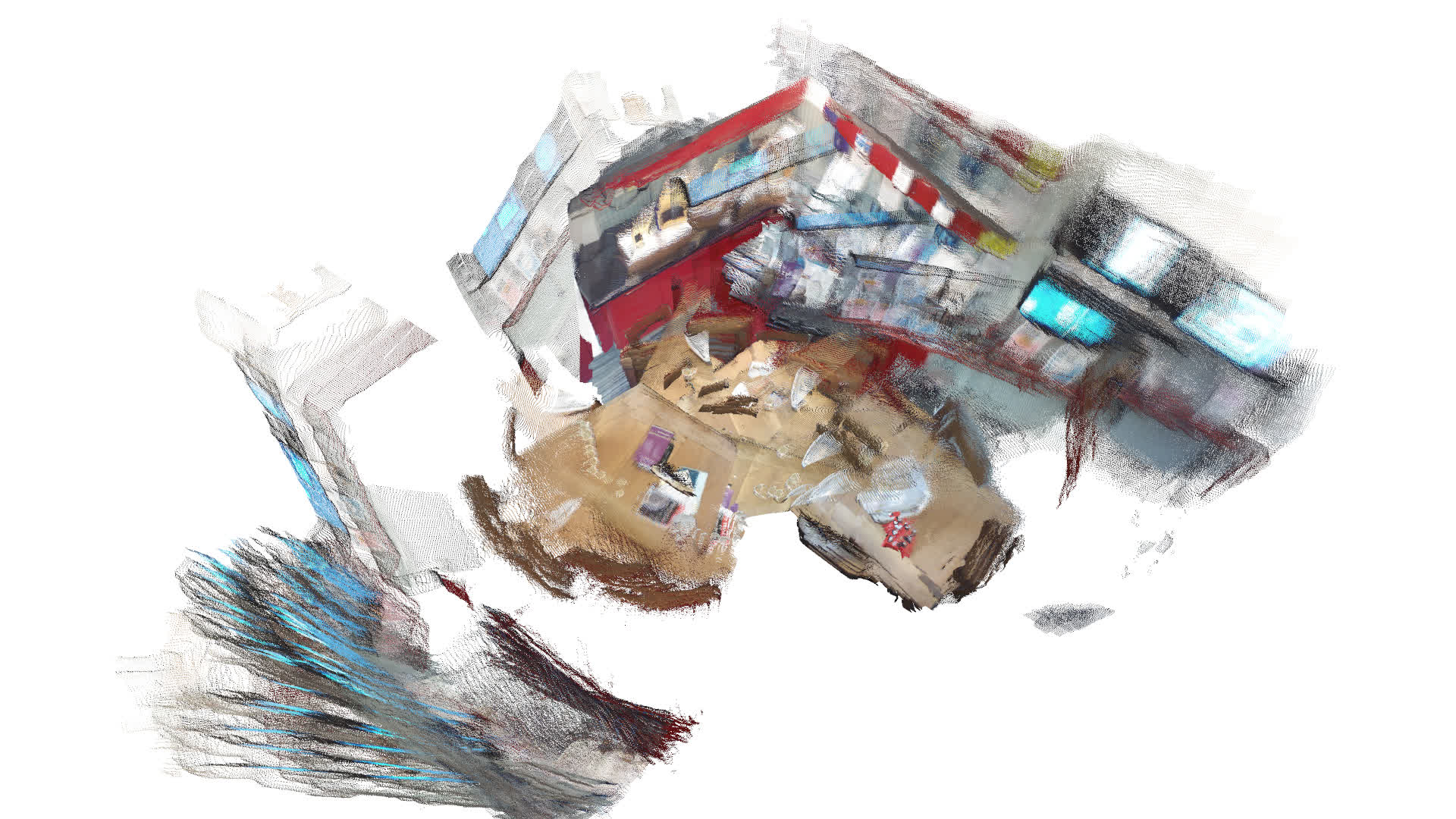} & \includegraphics[width=\sz\linewidth]{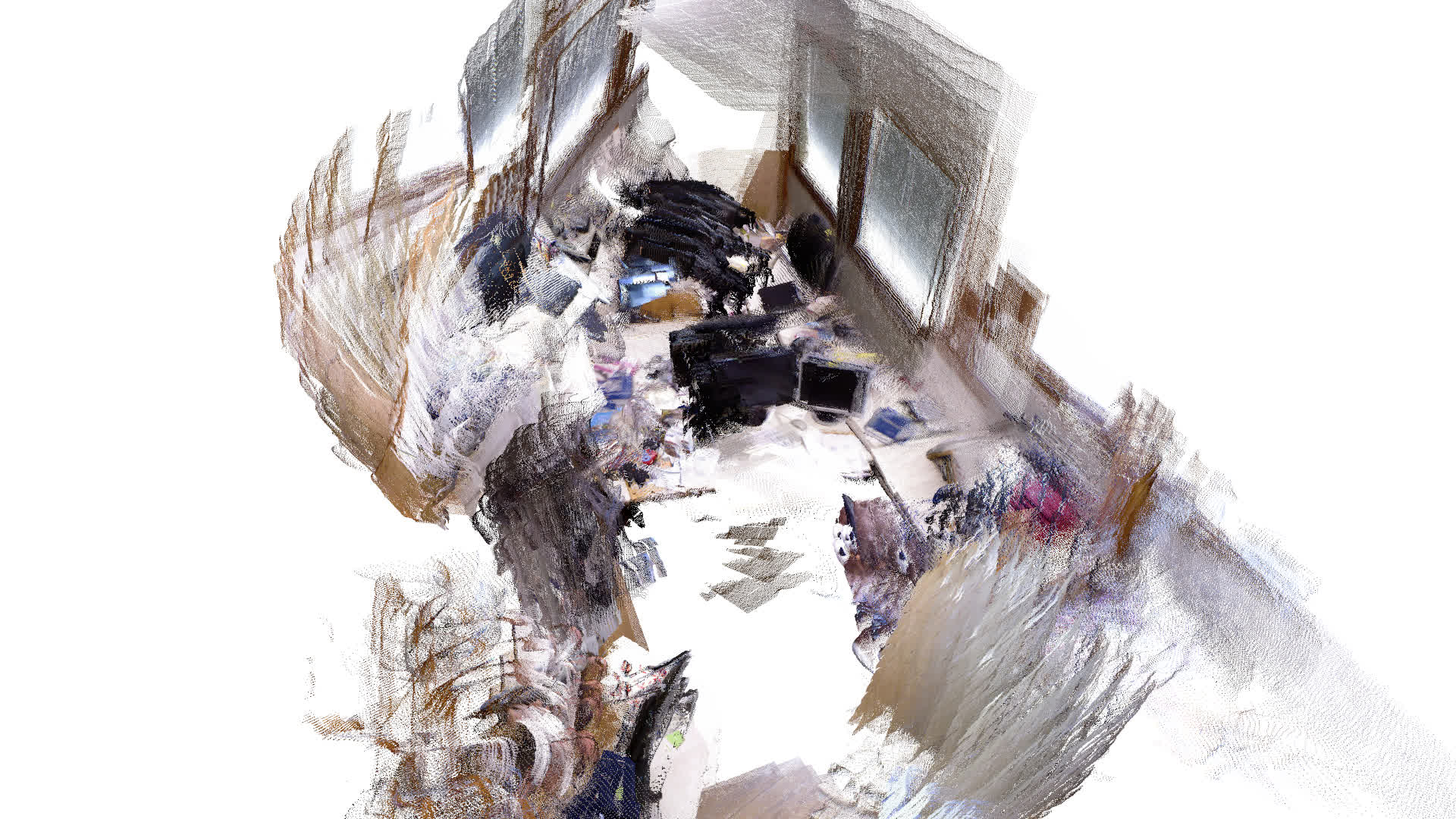} & 
        \includegraphics[width=\sz\linewidth]{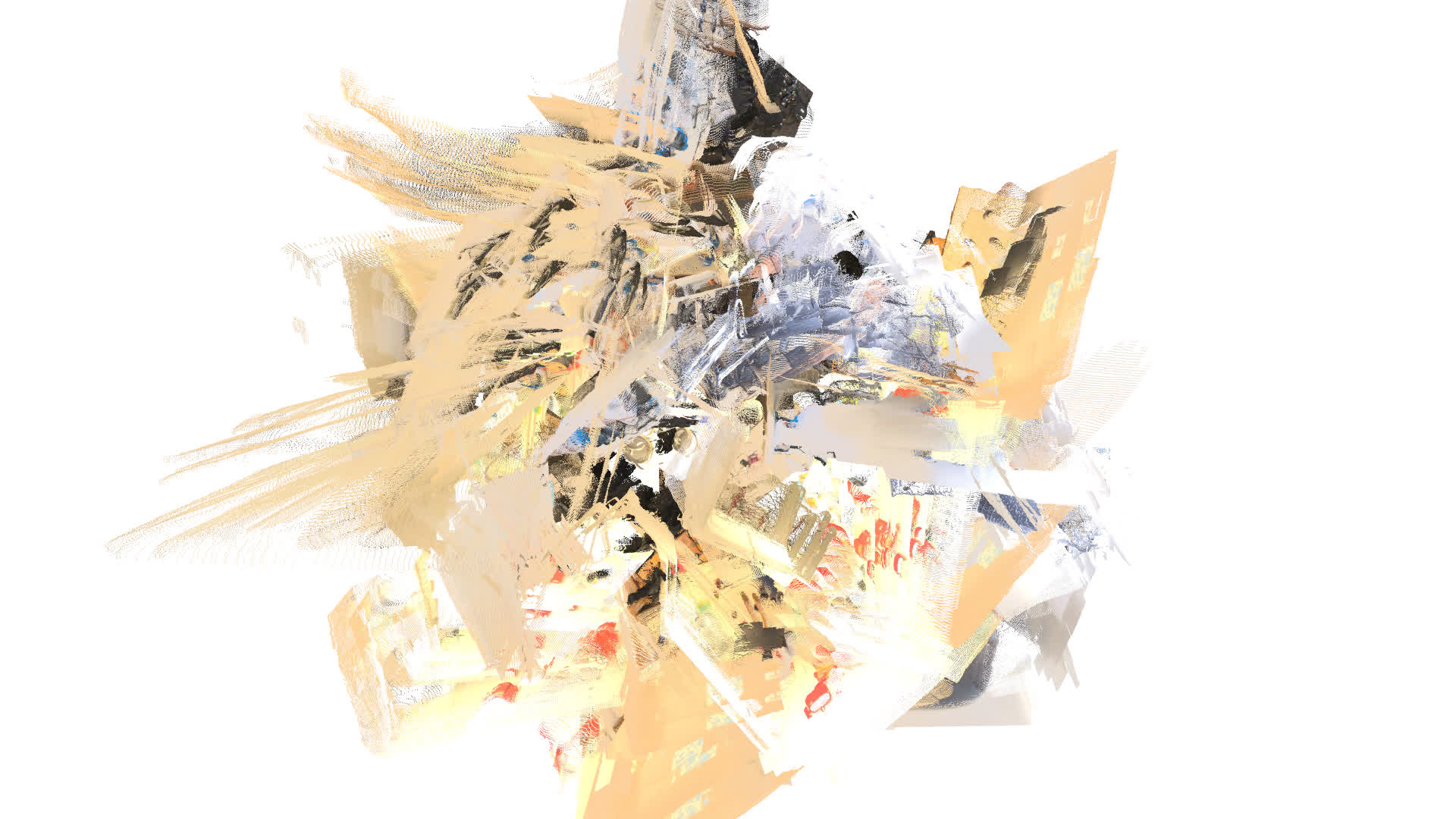}\\
    \rotatebox{90}{\hspace{1.2em} \small \makecell{SLAM3R~\cite{liu2025slam3r}}} &
        \includegraphics[width=\sz\linewidth]{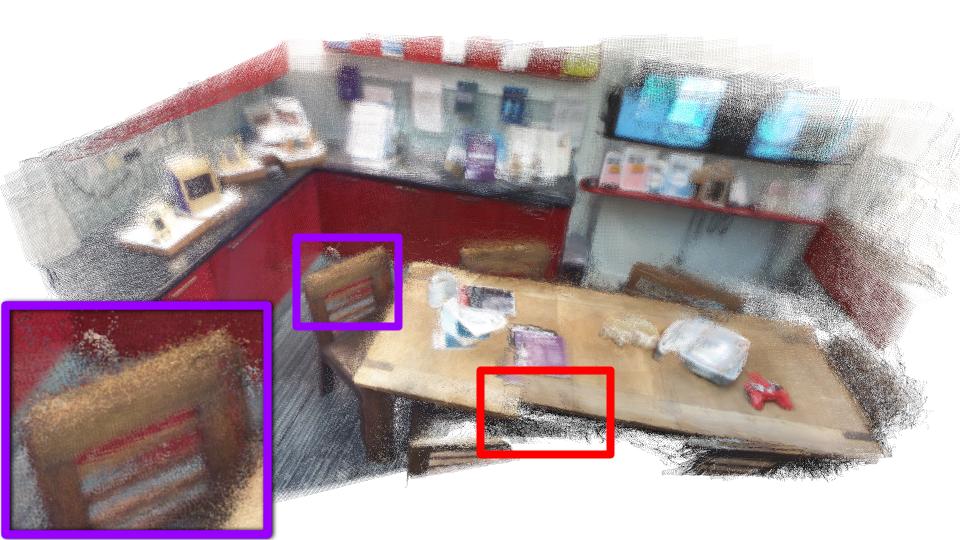} & \includegraphics[width=\sz\linewidth]{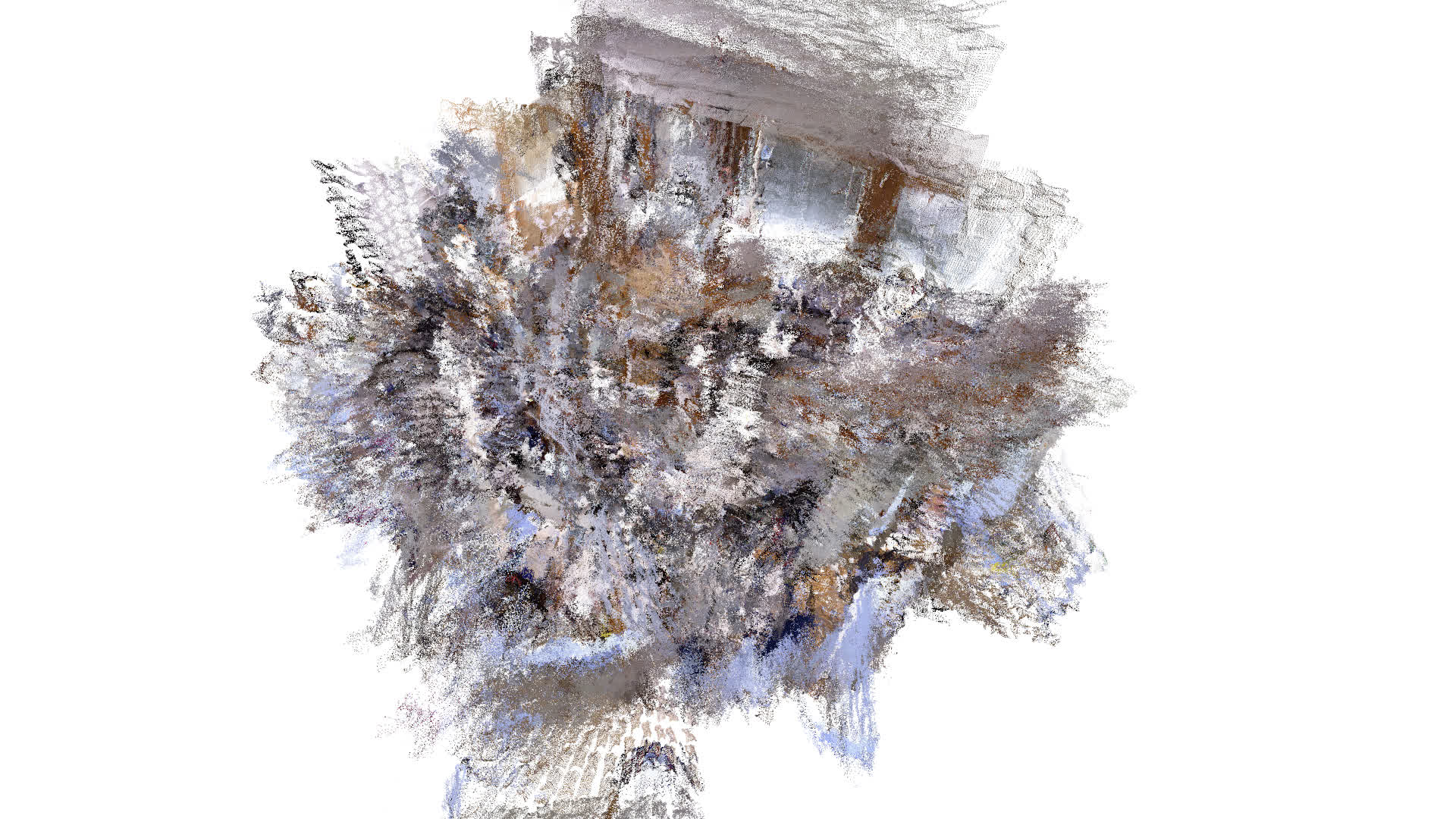} & 
        \includegraphics[width=\sz\linewidth]{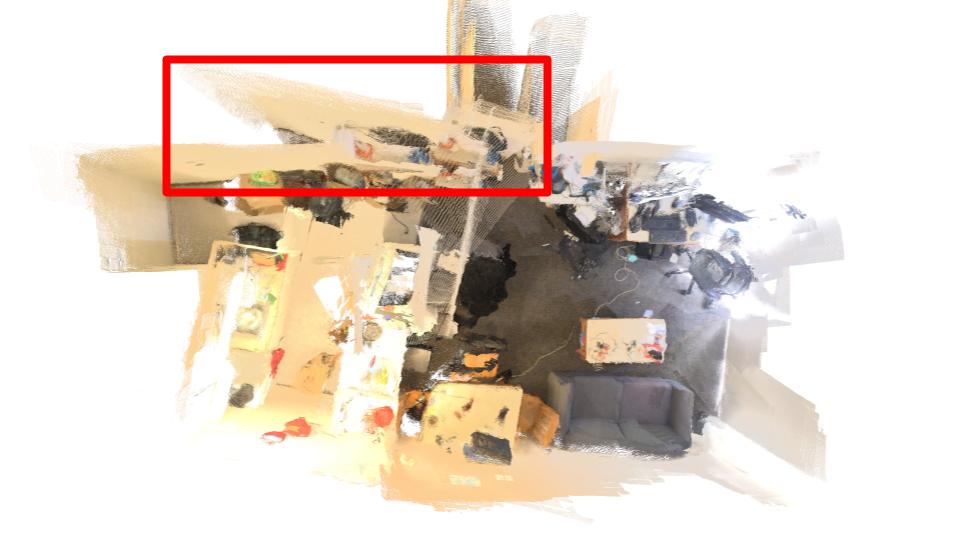}\\  
    \rotatebox{90}{\hspace{1.8em} \small \makecell{MASt3R-\\\hspace{0.5em}SLAM~\cite{murai2025mast3rslam}}} &
        \includegraphics[width=\sz\linewidth]{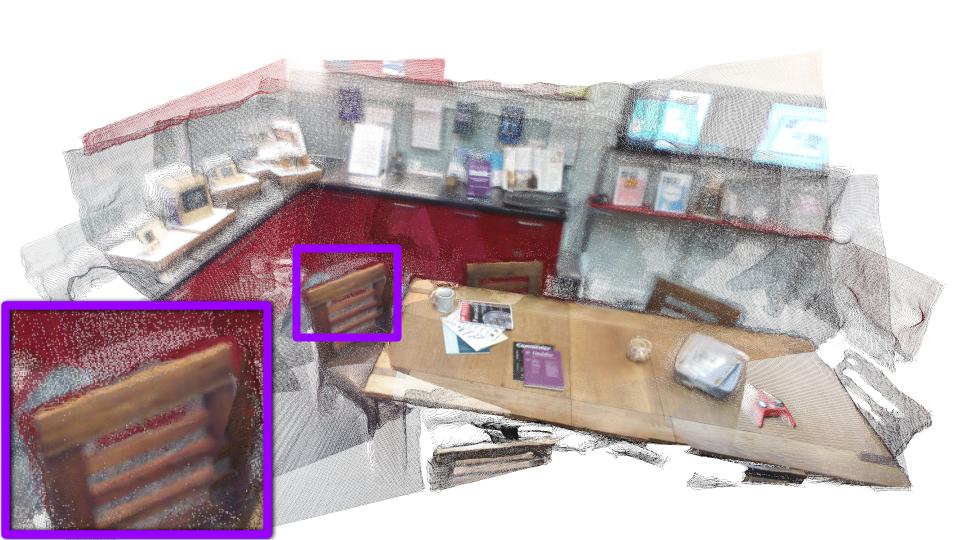} & \includegraphics[width=\sz\linewidth]{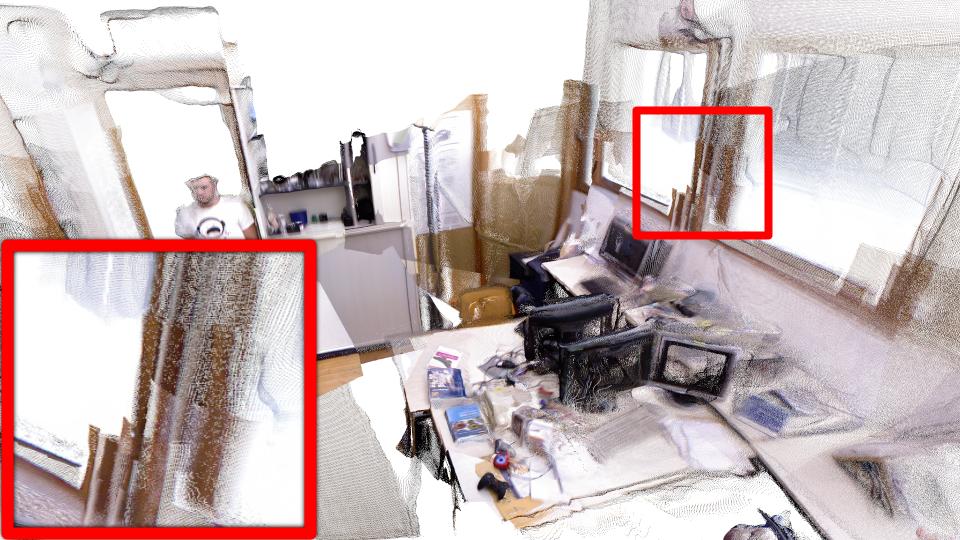} & 
        \includegraphics[width=\sz\linewidth]{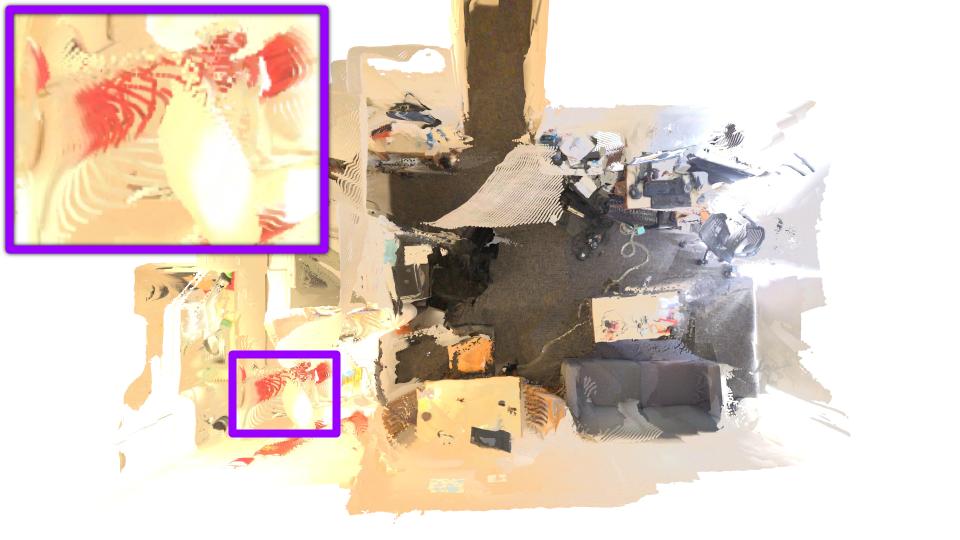}\\  
    \rotatebox{90}{\hspace{2.2em} \small \makecell{VGGT-\\\hspace{0.5em}SLAM~\cite{maggio2025vggtslam}}} &
        \includegraphics[width=\sz\linewidth]{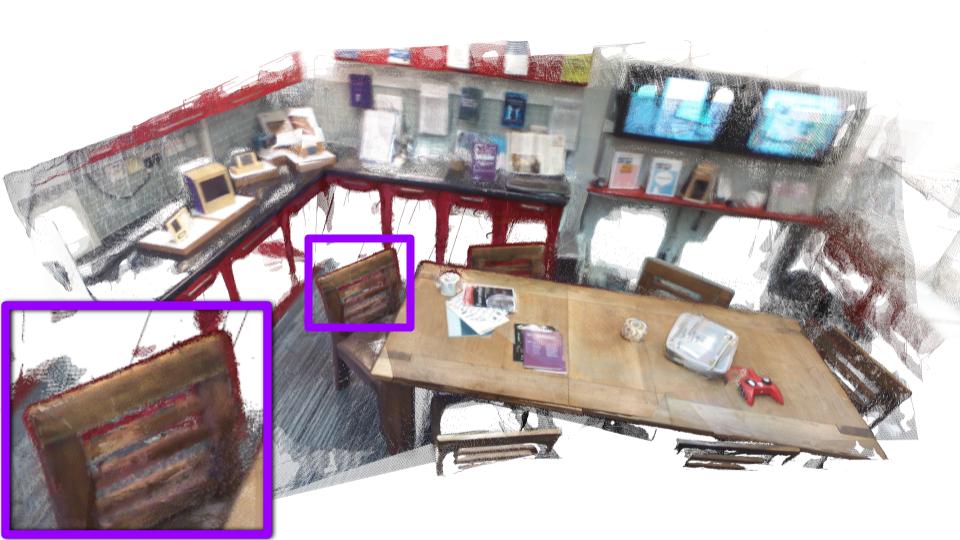} & \includegraphics[width=\sz\linewidth]{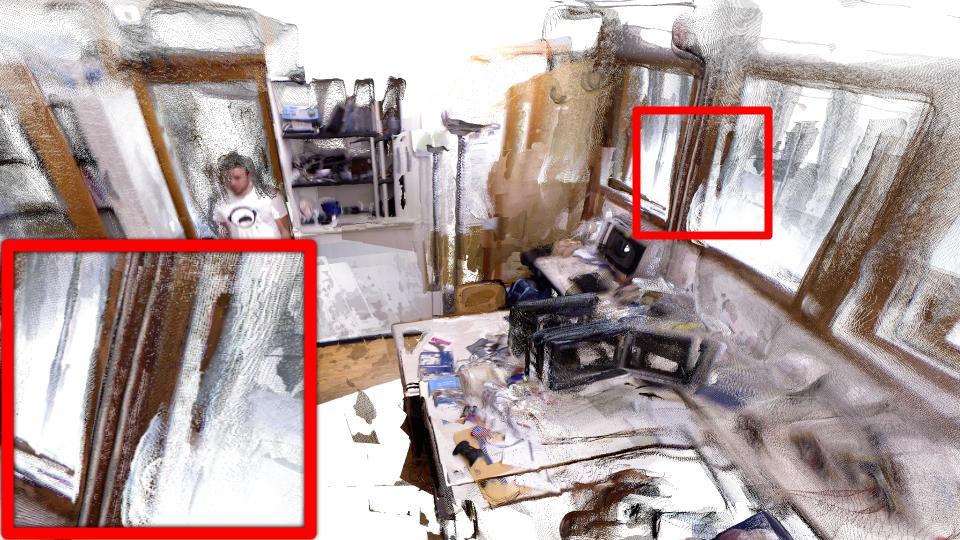} & 
        \includegraphics[width=\sz\linewidth]{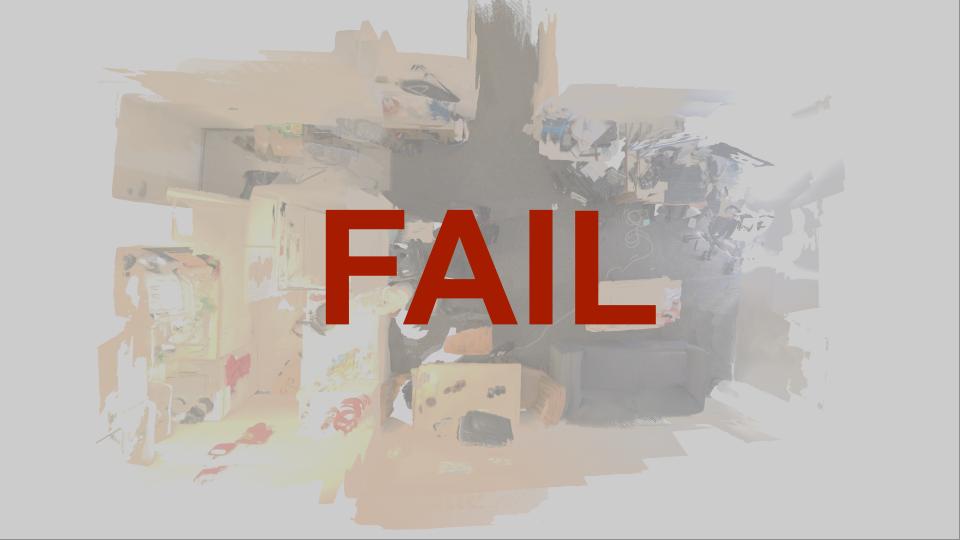}\\  
    \rotatebox{90}{\hspace{3.2em} \small \makecell{\bf \ours}} &
        \includegraphics[width=\sz\linewidth]{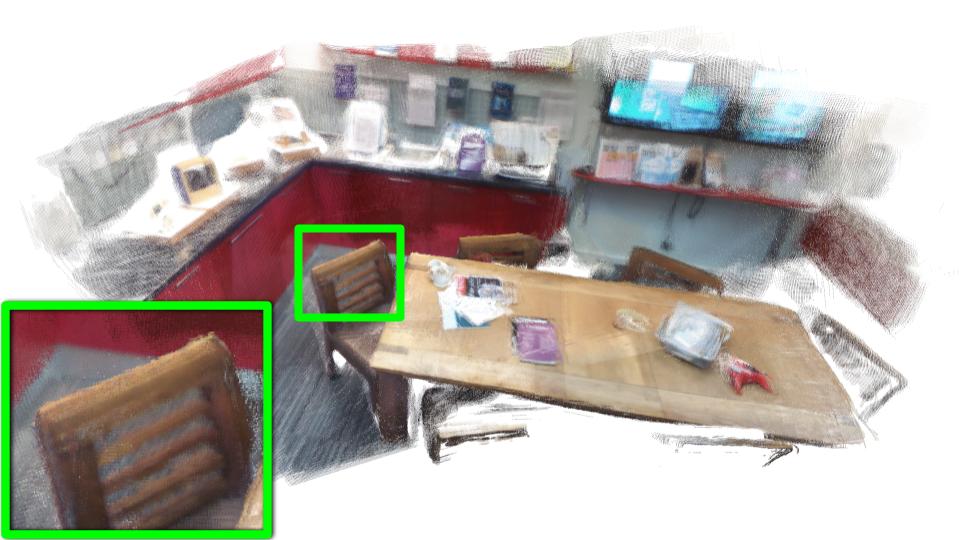} & \includegraphics[width=\sz\linewidth]{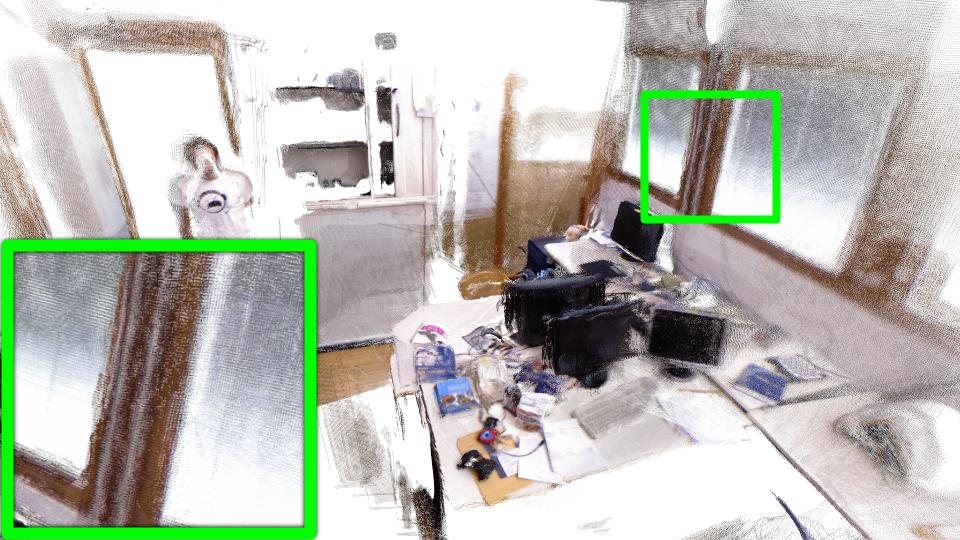} & 
        \includegraphics[width=\sz\linewidth]{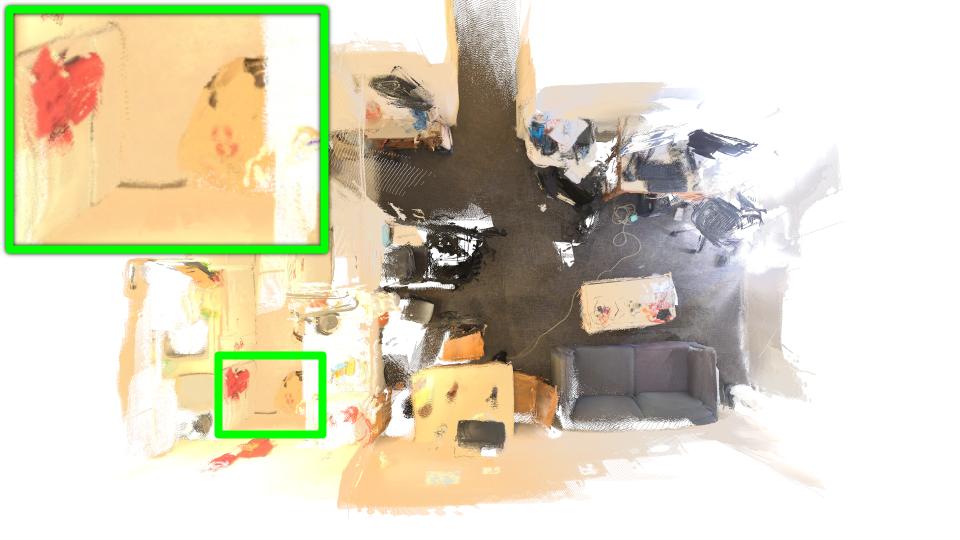}\\  
    \end{tabular}
}
\caption{\textbf{Reconstruction results on 7-Scenes \texttt{redkitchen} (left), TUM-RGBD \texttt{room} (middle), and BundleFusion \texttt{apt1} (right).} 
Purple boxes highlight reconstruction artifacts near the edges (background points wrongly mapped to the edge of the foreground). Red boxes indicate misalignments. Green boxes highlights \ours's competitive results.
VGGT-SLAM fails to complete reconstruction on \texttt{apt1} due to divergence in pose graph optimization.}
\label{fig:recon}
\vspace{-1em}
\end{figure*}

In \cref{fig:recon}, we show qualitative reconstruction results on 7-Scenes \texttt{redkitchen}, TUM-RGBD~\cite{sturm12tumrgbd} \texttt{room}, and BundleFusion~\cite{dai2017bundlefusion} \texttt{apt1}. CUT3R~\cite{wang2025cut3r} fails to reconstruct correctly due to forgetting issues, while SLAM3R~\cite{liu2025slam3r} struggles in scenes with large camera perspective changes. MASt3R-SLAM~\cite{murai2025mast3rslam} and VGGT-SLAM~\cite{maggio2025vggtslam} produce artifacts on object boundaries, failing to clearly separate foreground from background, and show misalignment across views. In contrast, \ours overcomes these challenges through geometric consistency constraints during training. Notably, VGGT-SLAM fails midway through the \texttt{apt1} scene as backend optimization diverges, which stems from the unstable RANSAC-based 3D homography estimation, which can sample planar regions and cause ambiguity in their proposed $\mathrm{SL}(4)$ pose graph optimization.

\subsection{Model Size and Speed}
\label{subsec:size_speed}

\begin{table}[t]
    \centering
    \small
    \setlength{\tabcolsep}{5pt}
    \resizebox{\columnwidth}{!}
    {
    \begin{tabular}{lcccccc}
    \toprule
      & load & encoder & decoder & detect & construct & optimize 
      \\ 
      & data &  &  & loop & graph & graph\\
    \midrule
    time (s) & 1.88 & 0.89 & 3.38 & 0.41 & 2.93 & 3.31\\
    \% & 14.7\% & 7.0\% & 26.4\% & 3.18\% & 22.9\% & 25.9\% \\
    \bottomrule
    \end{tabular}
    }
    \caption{
        \textbf{Time Spent on Each Component of \ours for 7-Scenes~\cite{shotton2013-7scenes} \texttt{redkitchen} (in seconds and percentage).}
    }
    \label{tab:step_time}
    \vspace{1em}

    \centering
    \small
    \setlength{\tabcolsep}{7.5pt}
    \begin{tabular}{llcc}
        \toprule
        \multicolumn{2}{c}{Settings}& ATE $\downarrow$ & Chamfer $\downarrow$ \\ 
        \midrule
        \multirow{2}{*}{\textit{STA Model}} 
        & \wo $L_\text{gc}$ & \nd 0.056 & \nd 0.057 \\
        & \wo $L_\text{id}$ & 0.058 & \rd 0.059 \\
        \midrule  
        \multirow{3}{*}{\textit{Pose Graph}} 
        & \wo pose graph opt. & 0.105 & 0.070 \\
        & \wo loop closure & 0.103 & 0.072 \\ 
        & \wo two edge types & \rd 0.057 & \st 0.051 \\
        
        \midrule
        \multicolumn{2}{c}{\bf \ours with full features} 
             & \st 0.055 & \st 0.051 \\
        \bottomrule
    \end{tabular}

    \caption{
        \textbf{Ablation Study on 7-Scenes~\cite{shotton2013-7scenes}.}  
        \textit{w/o pose graph opt.} simply accumulates relative poses for absolute poses. \textit{w/o two edge types} uses the classical pose graph in which each view is represented by a single node.
    }
    \label{tab:ablation}
    \vspace{-1.5em}
\end{table}

We compare the frontend model size and processing speed across methods in \cref{tab:size_and_time}. Owing to our symmetric design, the decoder and regression heads use only half the parameters of existing feedforward models~\cite{dust3r,mast3r,wang2025spann3r,liu2025slam3r}. Consequently, our model is far more compact: only 64\% the size of MASt3R~\cite{mast3r} (used in MASt3R-SLAM~\cite{murai2025mast3rslam}) and 35\% the size of VGGT~\cite{wang2025vggt} (used in VGGT-SLAM~\cite{maggio2025vggtslam}).

The speed evaluation further confirms that \ours achieves real-time performance. Benefiting from both the compact frontend and the sparse pose graph, our approach is highly competitive in runtime—faster than the pure regression-based methods CUT3R~\cite{wang2025cut3r} and SLAM3R~\cite{liu2025slam3r}, and comparable to VGGT-SLAM~\cite{maggio2025vggtslam}. It is worth noting that VGGT-SLAM performs inference only once every 32 keyframes, reducing the total number of inference steps. When replacing our STA model with a two-view VGGT that takes two views as input at a time like STA, the running speed is significantly slower, further demonstrating the effectiveness of our lightweight frontend.

\cref{tab:step_time} shows the percentage of runtime spent on major pipeline components. Decoding two-view information and pose graph optimization dominate the processing time.

\subsection{Ablation Study}
\label{subsec:ablation}

In \cref{tab:ablation}, we present ablations by selectively disabling components of \ours. Incorporating all proposed features yields the best performance on both camera trajectory estimation and 3D reconstruction.
 
Both $L_\text{gc}$ and $L_\text{id}$ contribute substantially to reconstruction quality. $L_\text{gc}$ improves consistency between the reconstructed local pointmaps of the two-view input pair, while $L_\text{id}$ further refines the estimated relative camera poses by enforcing a cycle consistency constraint on the model.


Pose graph optimization with loop closure is also highly effective, bringing 48\% improvement in the trajectory accuracy (0.105 $\rightarrow$ 0.055),  as it introduces simple yet powerful constraints through the edges of the pose graph, preventing error accumulation from two-view estimations as the trajectory grows longer. These findings further support the symmetric formulation of our frontend to regress local pointclouds and relative poses, which maximizes the effectiveness of pose graph optimization since the pointmaps of each view  are tightly coupled with their corresponding camera poses in the graph. On the contrary, if pointmaps were regressed in a shared coordinate system inside a submap, as in previous works~\cite{dust3r,mast3r,wang2025spann3r,cabon2025must3r,liu2025slam3r,wang2025cut3r}, pose updates would not be able to fix misalignments inside submaps.

Our two-edge-type pose graph design improves camera trajectory estimation by representing each view with multiple nodes connected by scale and pose edges, rather than a single node with standard edges. This structure better averages out uncertainty, particularly relative scale variations in pointmaps from different forward passes, improving the robustness of pose graph optimization and the accuracy of trajectory estimation.
\vspace{-0.2em}
\section{Conclusion}
\label{sec:conclusion}


We propose a novel monocular intrinsics-free SLAM pipeline, \ours, which features a lightweight frontend (Symmetric Two-View Association) and a $\mathrm{Sim}(3)$ pose graph optimization with loop closure as the backend. Experimental results demonstrate the superior camera tracking accuracy and 3D reconstruction quality of \ours. Meanwhile, it is significantly more lightweight and operates at a faster or comparable speed comparing current state-of-the-art methods.

\boldparagraph{Limitation and Future Work}
Our method omits optimization of point clouds in the backend for efficiency consideration. Therefore, it can suffer from misalignments caused by imperfect pointmap prediction by the frontend model. 
Future work could explore incorporating implicit camera information from previous estimates or aligning latent features across views to enhance local consistency across forward passes.
{   
    \clearpage
    \small
    \bibliographystyle{ieeenat_fullname}
    \bibliography{main}
}
\clearpage
\setcounter{page}{1}
\maketitlesupplementary

\begin{strip}
  \vspace{-0em}
  \centering
  \captionsetup{type=table} 

\renewcommand{\arraystretch}{1.1}
\setlength{\tabcolsep}{4.5pt}
\centering
\renewcommand{\arraystretch}{1.1}
\setlength{\tabcolsep}{4pt}
\begin{tabular}{llccccccccc}
\toprule
Method & Metric & \texttt{chess} & \texttt{fire} & \texttt{heads} & \texttt{office} & \texttt{pumpkin} & \texttt{redkitchen} & \texttt{stairs} & Avg \\
\midrule
\multirow{4}{*}{CUT3R} 
& Accuracy     & 0.274 & 0.102 & 0.106 & 0.326 & 0.452 & 0.325 & 0.345 & 0.276 \\
& Completeness & 0.303 & 0.081 & 0.093 & 0.441 & 0.459 & 0.358 & 0.389 & 0.303 \\
& Chamfer      & 0.289 & 0.091 & 0.100 & 0.384 & 0.456 & 0.342 & 0.367 & 0.290 \\
& ATE   & 0.743 & 0.226 & 0.363 & 0.664 & 0.546 & 0.381 & 0.413 & 0.477 \\
\midrule
\multirow{4}{*}{SLAM3R} 
& Accuracy     & 0.043 & 0.022 & 0.020 & 0.035 & 0.072 & 0.062 & 0.116 & 0.053 \\
& Completeness & 0.030 & 0.013 & 0.015 & 0.030 & 0.055 & 0.061 & 0.209 & \underline{0.059} \\
& Chamfer      & 0.037 & 0.018 & 0.017 & 0.033 & 0.064 & 0.061 & 0.162 & \underline{0.056} \\
& ATE   & 0.089 & 0.048 & 0.036 & 0.088 & 0.196 & 0.102 & 0.126 & 0.098 \\
\midrule
\multirow{4}{*}{MASt3R-SLAM} 
& Accuracy     & 0.090 & 0.037 & 0.027 & 0.047 & 0.097 & 0.070 & 0.045 & 0.059 \\
& Completeness & 0.055 & 0.024 & 0.021 & 0.053 & 0.054 & 0.036 & 0.149 & \bf{0.056} \\
& Chamfer      & 0.073 & 0.031 & 0.024 & 0.050 & 0.075 & 0.053 & 0.097 & 0.057 \\
& ATE   & 0.063 & 0.046 & 0.029 & 0.103 & 0.112 & 0.074 & 0.032 & \underline{0.066} \\
\midrule
\multirow{4}{*}{VGGT-SLAM} 
& Accuracy     & 0.029 & 0.014 & 0.031 & 0.041 & 0.128 & 0.036 & 0.087 & \underline{0.052} \\
& Completeness & 0.052 & 0.064 & 0.021 & 0.066 & 0.054 & 0.057 & 0.110 & 0.061 \\
& Chamfer      & 0.040 & 0.039 & 0.026 & 0.054 & 0.091 & 0.047 & 0.098 & \underline{0.056} \\
& ATE   & 0.037 & 0.026 & 0.022 & 0.103 & 0.147 & 0.063 & 0.095 & 0.070 \\
\midrule
\multirow{4}{*}{Ours} 
& Accuracy     & 0.065 & 0.015 & 0.031 & 0.036 & 0.061 & 0.035 & 0.074 & \bf{0.045} \\
& Completeness & 0.063 & 0.022 & 0.040 & 0.048 & 0.037 & 0.030 & 0.154 & \bf{0.056} \\
& Chamfer      & 0.064 & 0.019 & 0.035 & 0.042 & 0.049 & 0.033 & 0.114 & \bf{0.051} \\
& ATE   & 0.073 & 0.035 & 0.028 & 0.055 & 0.129 & 0.035 & 0.029 & \bf 0.055 \\
\bottomrule
\end{tabular}
\vspace{7pt}
\caption{\textbf{Per Scene Evaluation on 7-Scenes~\cite{shotton2013-7scenes}.} Comparison of accuracy, completeness, Chamfer distance, and trajectory error on the 7-Scenes dataset. Lower is better. Best results are \textbf{bold}, second best are \underline{underlined}.}
\label{tab:7scenes_results_full}
\vspace{4em}
\end{strip}

\section{Relative Scale in Pose Graph}
As mentioned in \cref{subsec:backend}, the \textit{scale edges} connect nodes corresponding to the same view but obtained from different forwarding passes. Since the training supervision of the frontend STA model uses only normalized pointmaps, the scales of the same view across different passes are not consistent. Therefore, estimating the relative scale is crucial for pose graph construction.
Given two pointmaps $\bm{P}_i^j$ and $\bm{P}_i^k$ of the same view $i$ (obtained from forwarding passes with input view $i\  j$ and view $i\ k$, respectively), along with their confidence maps $\bm{C}_i^j$ and $\bm{C}_i^k$, we first get the confidence score $w_{x}$ of the point pair for pixel $\bm x$,
\begin{align}
    w_{x} = &\bm C_i^j(\bm x) \cdot \bm C_i^k(\bm x), \nonumber
\end{align}
then, the relative scale $s_i^{jk}$ can be computed as,
\begin{align}
    s_i^{jk} =& \min_s \sum_{\bm x} w_{x}\| \bm P_i^j(\bm x)-s\bm P_i^k(\bm x) \|^2 \nonumber \\
    =& \dfrac{\sum_{\bm x}w_{x}(\bm P_i^j(\bm x)\cdot \bm P_i^k(\bm x))}{\sum_{\bm x}w_{x}\|\bm P_i^k(\bm x)\|^2}.
\end{align}

\section{Additional Quantitative Results}

\subsection{Per Scene Evaluation Results on 7-Scenes}
In \cref{subsec:recon}, only the average reconstruction evaluation is provided. In \cref{tab:7scenes_results_full}, we present more detailed per-scene results on 7-Scenes~\cite{shotton2013-7scenes} to offer deeper insights. The pure regression method SLAM3R~\cite{liu2025slam3r} performs well in scenes where the camera primarily focuses on a single corner, such as \texttt{chess}, \texttt{fire}, and \texttt{heads}. However, in scenes involving longer camera trajectories, like \texttt{pumpkin} and \texttt{redkitchen}, its performance degrades due to difficulties in accurately registering points. Our method, \ours, achieves the best performance in average across all four metrics.

\subsection{Additional Trajectory Results on TUM-RGBD}
In \cref{subsec:traj}, to align with previous methods, only results for the \textit{freiburg1} partition of the TUM RGB-D dataset~\cite{sturm12tumrgbd} are reported. In \cref{tab:traj_tumrgbd_more}, we also report results for the \textit{freiburg2} and \textit{freiburg3} partitions.

\begin{table}[h]
    \centering
    \begin{tabular}{l c}
        \toprule
        \textbf{Sequence} & \textbf{ATE RMSE (m)} \\
        \midrule
        freiburg2\_360\_hemisphere        & 0.2037 \\
        freiburg2\_360\_kidnap            & 0.4617 \\
        freiburg2\_desk                   & 0.0577 \\
        freiburg2\_large\_with\_loop      & 0.2170 \\
        freiburg2\_rpy                    & 0.0222 \\
        freiburg2\_xyz                    & 0.0155 \\
        \midrule
        freiburg3\_cabinet                & 0.3869 \\
        freiburg3\_large\_cabinet         & 0.1334 \\
        freiburg3\_long\_office\_household & 0.1013 \\
        freiburg3\_teddy                  & 0.0789 \\
        \bottomrule
    \end{tabular}
    \caption{\bf Trajectory ATE results on the \textit{freiburg2} and \textit{freiburg3} partitions of the TUM RGB-D dataset~\cite{sturm12tumrgbd}.}
        \label{tab:traj_tumrgbd_more}
\end{table}

\subsection{Additional Evaluation on More Datasets}
In \cref{tab:replica_eval} and \cref{tab:scannet_eval}, we additionally report camera trajectory evaluation results on Replica~\cite{replica19arxiv} and ScanNet~\cite{dai2017scannet}, as well as reconstruction evaluation results on Replica for several commonly used SLAM testing scenes.

\begin{table}[h]
    \centering
    \begin{tabular}{l c c c c}
        \toprule
        \textbf{Scene} & \textbf{ATE} & \textbf{Acc.} & \textbf{Comp.} & \textbf{Chamfer} \\
        \midrule
        office0 & 0.0744 & 0.0595 & 0.0226 & 0.0410 \\
        office1 & 0.1934 & 0.2614 & 0.1833 & 0.2223 \\
        office2 & 0.1177 & 0.0914 & 0.0316 & 0.0615 \\
        office3 & 0.0485 & 0.0623 & 0.0221 & 0.0422 \\
        office4 & 0.1302 & 0.1338 & 0.0688 & 0.1013 \\
        room0   & 0.0688 & 0.0766 & 0.0209 & 0.0488 \\
        room1   & 0.0934 & 0.1105 & 0.0552 & 0.0828 \\
        room2   & 0.1363 & 0.1194 & 0.0223 & 0.0709 \\
        \bottomrule
    \end{tabular}
    \caption{\bf Per-scene evaluation results on Replica~\cite{replica19arxiv}.}
    \label{tab:replica_eval}
    \vspace{1.5em}

\setlength{\tabcolsep}{25pt}
    \centering
    \begin{tabular}{l c}
        \toprule
        \textbf{Sequence} & \textbf{ATE RMSE (m)} \\
        \midrule
        scene0000\_00 & 0.0483 \\
        scene0059\_00 & 0.0391 \\
        scene0106\_00 & 0.0559 \\
        scene0169\_00 & 0.0526 \\
        scene0181\_00 & 0.0520 \\
        scene0207\_00 & 0.0479 \\
        \bottomrule
    \end{tabular}
    \caption{\bf Per-scene camera trajectory evaluation results on ScanNet~\cite{dai2017scannet}.}
        \label{tab:scannet_eval}
\end{table}

\begin{figure*}[!htb]
\centering
{\small
    \setlength{\tabcolsep}{15pt}
    \renewcommand{\arraystretch}{1}
    \newcommand{\sz}{0.4}
    \begin{tabular}{cc}
      \includegraphics[width=\sz\linewidth]{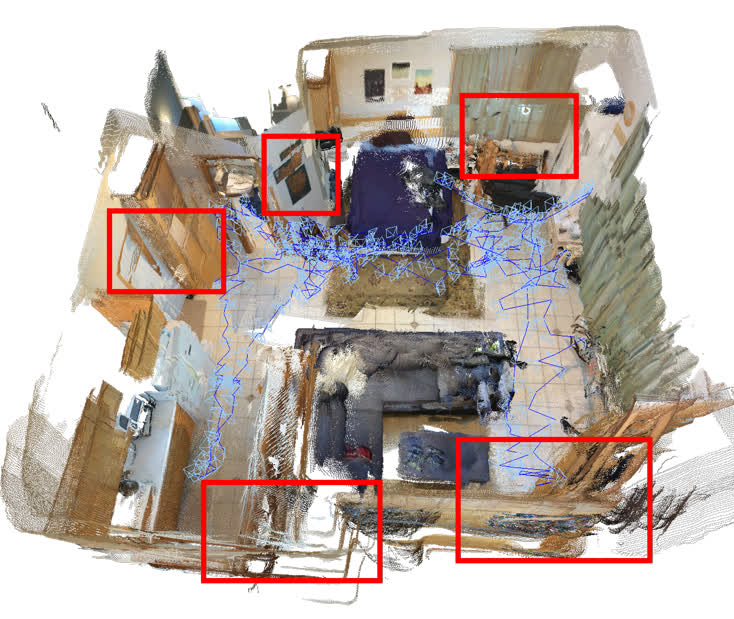} & 
      \includegraphics[width=\sz\linewidth]{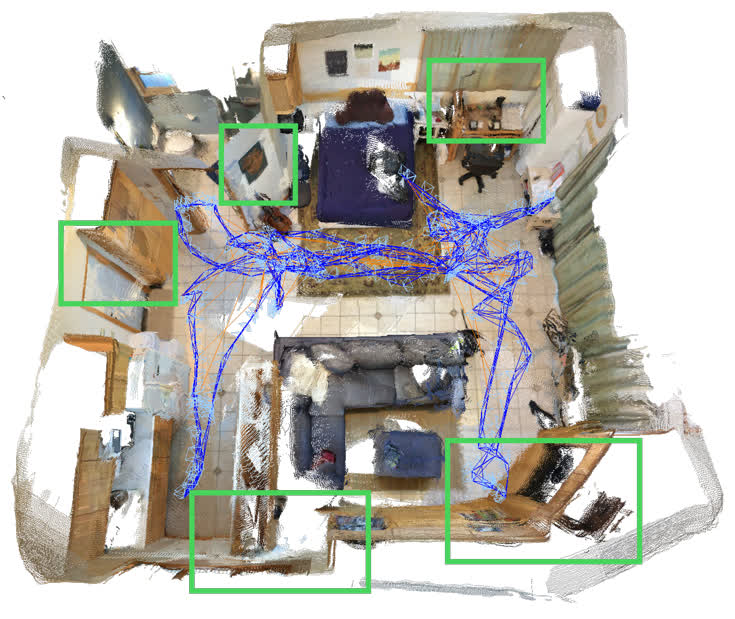} \\ 
      result \wo PGO & result \emph{w/} PGO

    \end{tabular}
}
\vspace{7pt}
\caption{\textbf{Qualitative Comparison for Pose Graph Optimization.} Red boxes highlight regions with misalignments, while green boxes indicate areas where these misalignments have been corrected after pose graph optimization.
}
\label{fig:pgo}
\vspace{2.5em}

\centering
\includegraphics[width=\linewidth]{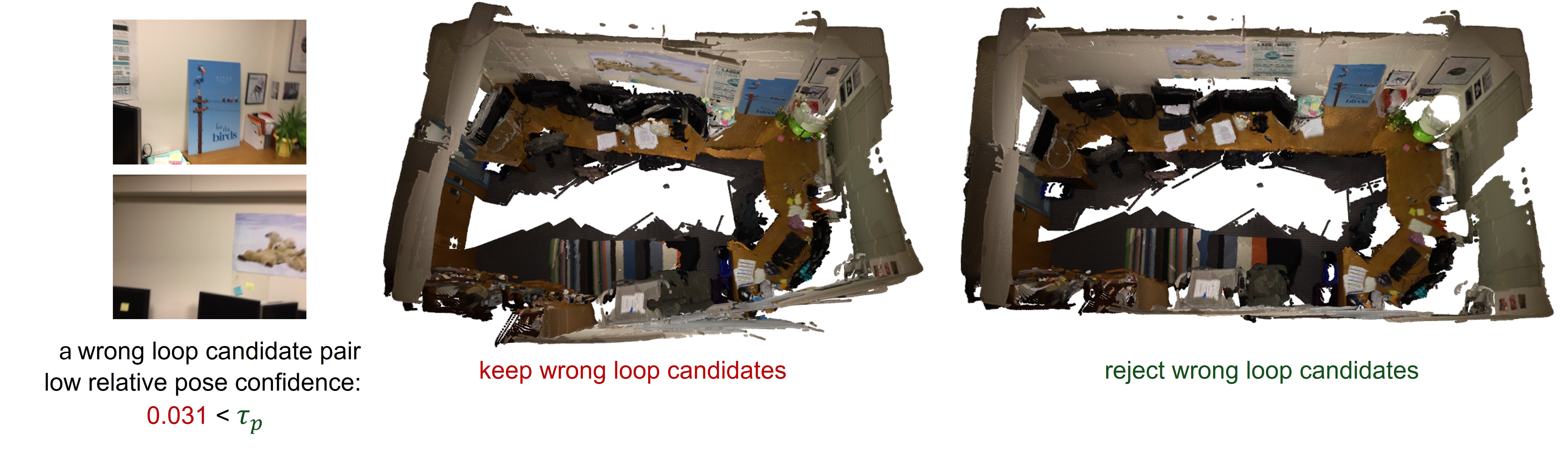}
\vspace{-1.7em}
\caption{\textbf{Qualitative Comparison of Wrong Loop Filtering.} Keeping wrong loop candidates decreases the performance a lot.}
\label{fig:wrong_loop}

\end{figure*}

\section{Additional Qualitative Results}

\subsection{Pose Graph Optimization}
In \cref{fig:pgo}, we compare the reconstruction and trajectory estimation results with and without pose graph optimization on ScanNet~\cite{dai2017scannet} \texttt{scene0000\_00}. Pose graph optimization effectively corrects misaligned areas and averages out the errors from the frontend.

\subsection{Wrong Loop Filtering}
In \cref{subsec:backend}, we describe feeding each loop candidate pair into our STA model to verify their spatial proximity. This is necessary because Bag of Words loop detection can produce false positives, which may significantly degrade performance by introducing misleading edges into the pose graph. As shown in \cref{fig:wrong_loop}, rejecting incorrect loop candidates using the relative pose confidence score provided by STA results in a much more stable performance.

\subsection{More Results}
\begin{figure*}[!htb]
\centering
{\small
    \setlength{\tabcolsep}{25pt}
    \renewcommand{\arraystretch}{1}
    \newcommand{\sz}{0.38}
    \begin{tabular}{cc}
      BundleFusion~\cite{dai2017bundlefusion} \texttt{apt2} & TUM-RGBD~\cite{sturm12tumrgbd} \texttt{floor} \\
      \includegraphics[width=\sz\linewidth]{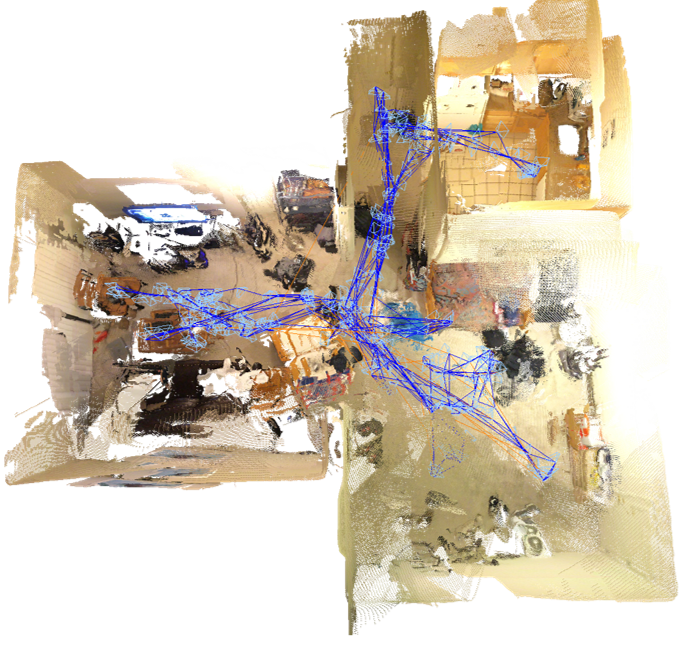} & 
      \includegraphics[width=\sz\linewidth]{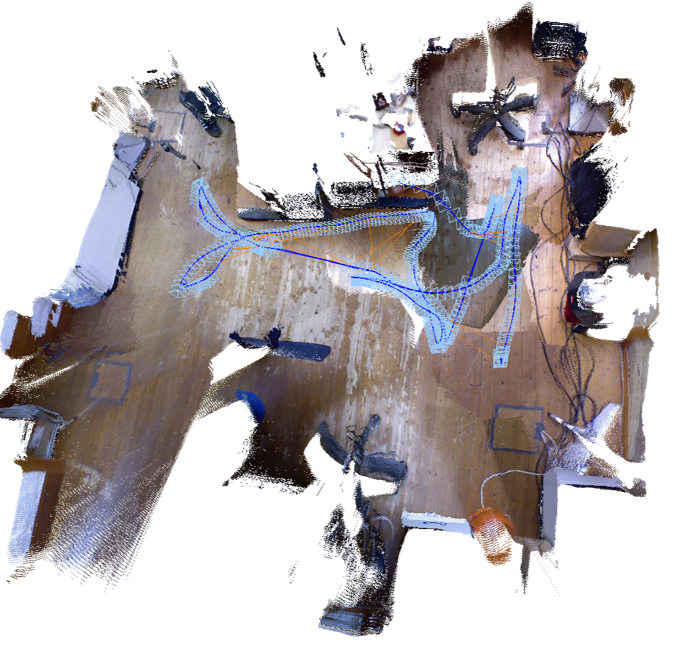} \\ 
      
      BundleFusion~\cite{dai2017bundlefusion} \texttt{apt0} & 7-Scenes~\cite{shotton2013-7scenes} \texttt{pumpkin}  \\
      \includegraphics[width=\sz\linewidth]{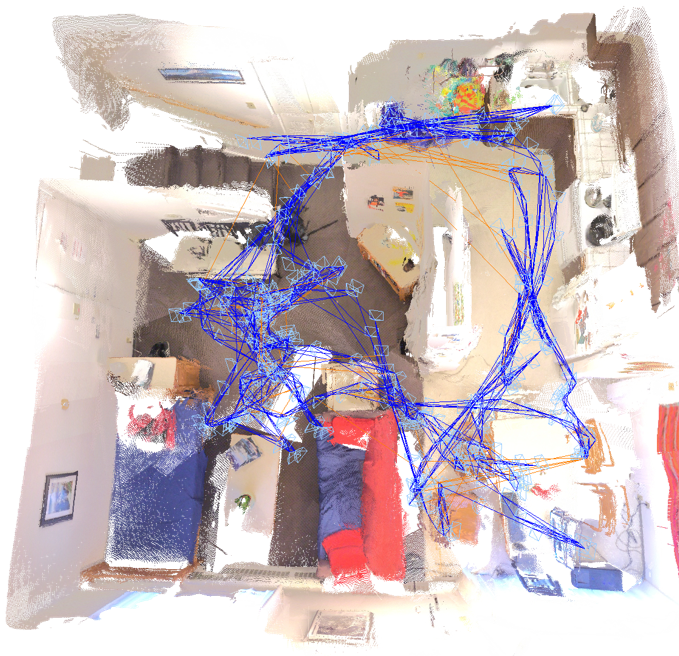} & 
      \includegraphics[width=\sz\linewidth]{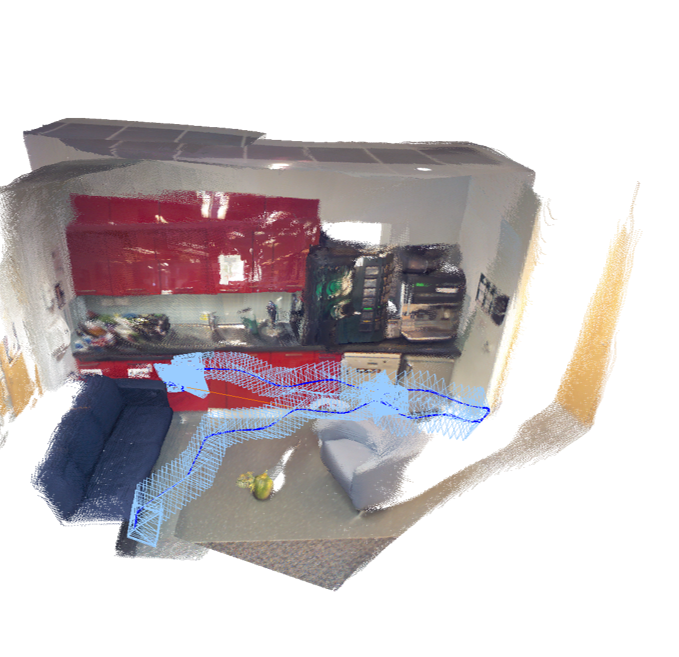} \\ 

      M2DGR~\cite{yin2021m2dgr} \texttt{room\_03} & M2DGR~\cite{yin2021m2dgr} \texttt{room\_01}  \\
      \includegraphics[width=\sz\linewidth]{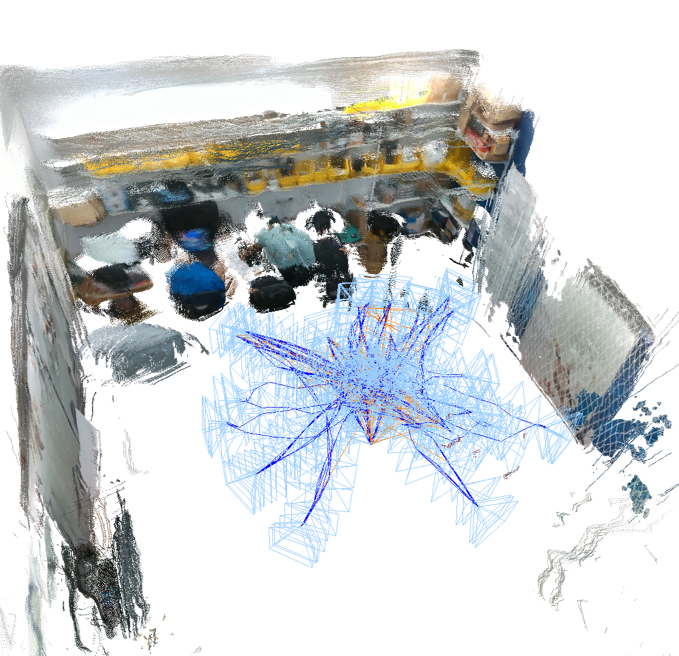} & 
      \includegraphics[width=\sz\linewidth]{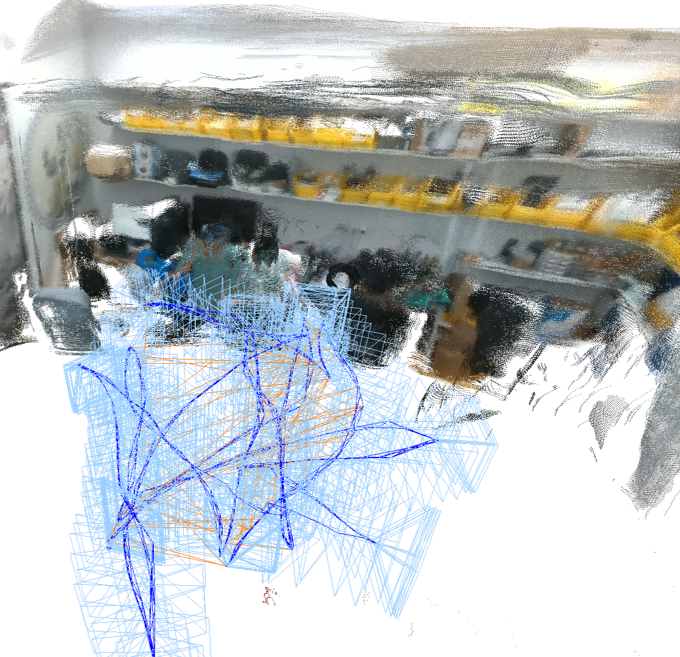} \\ 
    \end{tabular}
}
\caption{\textbf{More Qualitative Results.} Reconstructions and camera trajectories from different datasets.}
\label{fig:more}
\end{figure*}

In \cref{fig:more}, we present the results of \ours across various datasets. \ours demonstrates stable performance despite differing camera motions in these scenes. As before, light blue frustums represent camera poses, blue lines connect neighboring views, while orange lines indicate loop closures.

\end{document}